%% file: coling_latex.tex
\pdfoutput=1

\documentclass[11pt]{article}

\usepackage[preprint]{coling}

\usepackage{times}
\usepackage{latexsym}

\usepackage{graphicx,xcolor} 
\usepackage{booktabs,multirow}
\usepackage{amsmath,amssymb,mathtools}
\usepackage{color}
\usepackage{lingmacros}
\usepackage{graphbox}
\usepackage{colortbl}
\usepackage{url}
\usepackage{tikz}
\usepackage{tikz-dependency}
\usetikzlibrary{positioning} 
\usepackage {diagbox}
\usepackage{subcaption} %
\usepackage{amssymb}
\usepackage{xspace}

\usepackage{cleveref}
\Crefname{section}{\S}{\S\S}

\newcommand{\gcella}{\cellcolor[rgb]{0.9, 0.95, 0.9}}
\newcommand{\gcellb}{\cellcolor[rgb]{0.8, 0.95, 0.8}}
\newcommand{\gcellc}{\cellcolor[rgb]{0.65,0.95, 0.65}}

\newcommand{\rcella}{\cellcolor[rgb]{0.95, 0.9, 0.9}}
\newcommand{\rcellb}{\cellcolor[rgb]{0.95, 0.8, 0.8}}
\newcommand{\rcellc}{\cellcolor[rgb]{0.95, 0.7, 0.7}}
\newcommand{\grcell}{\cellcolor[rgb]{0.9, 0.9, 0.9}}

\newcommand{\gl}[2]{%
\leavevmode\vtop{\hbox{#1}%
\hbox{#2\lower1.4ex\rlap{ }}}}

\definecolor{mycyan}{rgb}{0.8, 1, 1}
\newcommand{\sight}{\includegraphics[align=c, width=1.2em]{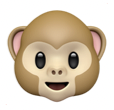}\xspace}
\newcommand{\blind}{\includegraphics[align=c, width=1.2em]{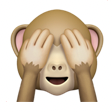}\xspace}
\newcommand{\cyclone}{\includegraphics[align=c, width=1.2em]{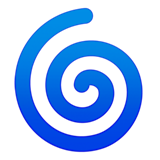}\xspace}

\newcommand{\sighted}{\checkmark}
\newcommand{\diff}{$\Delta$}
\newcommand{\mathboldface}[1]{\boldsymbol{#1}}
\newcommand{\bm}[1]{\mathboldface{#1}}

\usepackage[T1]{fontenc}

\usepackage[utf8]{inputenc}

\usepackage{microtype}

\usepackage{inconsolata}

\usepackage{graphicx}

\title{Does Vision Accelerate Hierarchical Generalization in \\ Neural Language Learners?}

\author{Tatsuki Kuribayashi \and Timothy Baldwin
  \\
  MBZUAI
  \\
  \texttt{tatsuki.kuribayashi@mbzuai.ac.ae}
}

\date{}

\begin{document}
\maketitle
\begin{abstract}
Neural language models (LMs) are arguably less data-efficient than humans from a language acquisition perspective.
One fundamental question is why this human--LM gap arises. 
This study explores the advantage of \textit{grounded} language
acquisition, specifically the impact of visual information --- which humans can usually rely on but LMs largely do not have access to during language acquisition --- on syntactic generalization in LMs.
Our experiments, following the poverty of stimulus paradigm under two scenarios (using artificial vs.\ naturalistic images), demonstrate that if the alignments between the linguistic and visual components are clear in the input, access to vision data does help with the syntactic generalization of LMs, but if not, visual input does not help.
This highlights the need for additional biases or signals, such as mutual gaze, to enhance cross-modal alignment and enable efficient syntactic generalization in multimodal LMs.

\end{abstract}

\section{Introduction}
\label{sec:introduction}

Neural language models (LMs) have accelerated progress in natural language processing (NLP), but there remains a significant disparity in their data efficiency compared to humans. 
For instance, GPT-3~\cite{Brown2020-zt} is trained on approximately 2,000 times more text than a 10-year-old child is exposed to~\cite{Warstadt2022-zi} and this gap is even greater in modern large LMs, and yet the model still struggles with some language tasks.
We investigate what kind of differences between human and LM language acquisition scenarios can potentially close the gap in data efficiency, specifically to achieve syntactic generalization.

\begin{figure}[t]
    \centering
      \includegraphics[width=\linewidth]{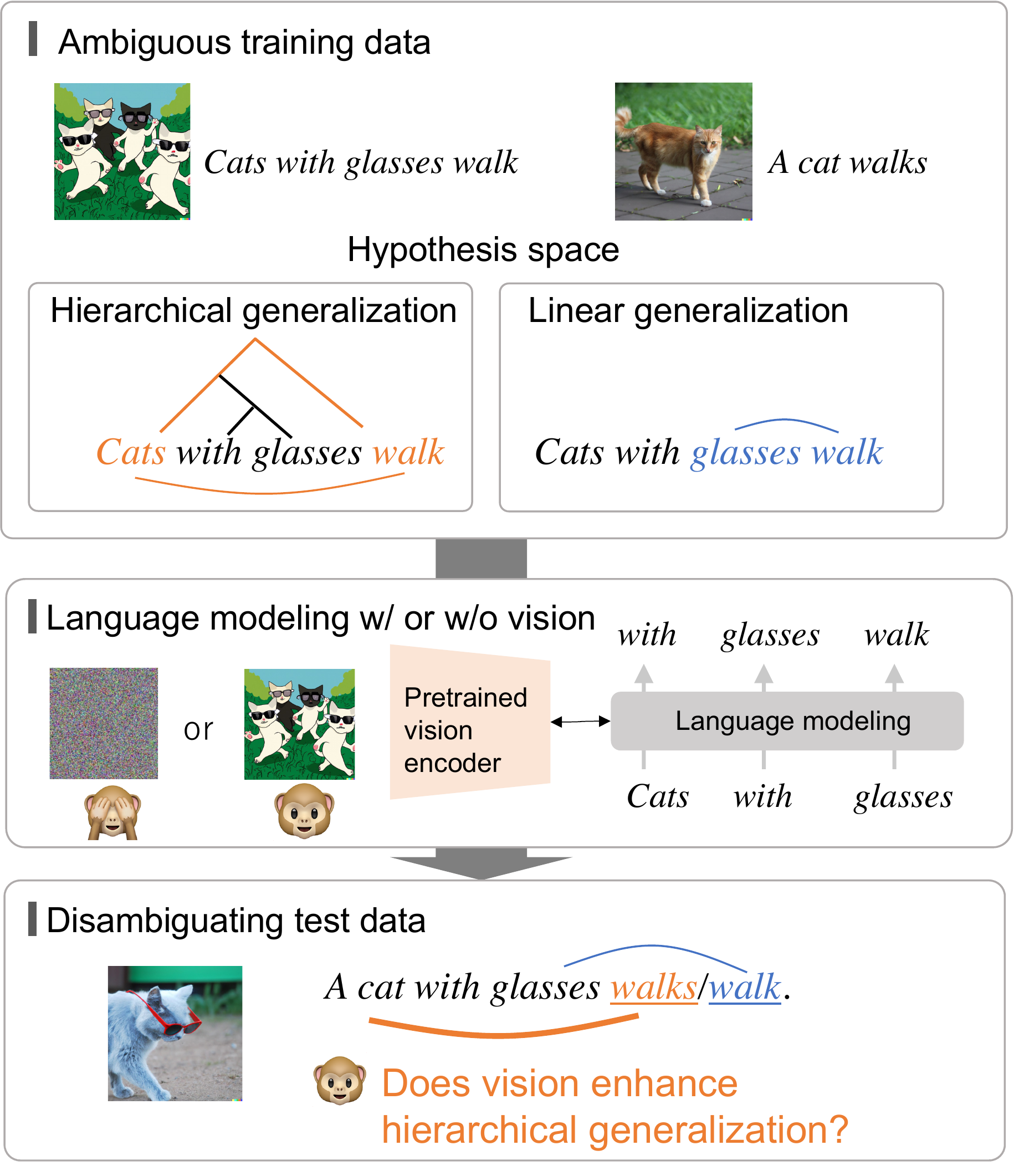}
      \caption{Overview of the experimental design. A vision-language neural model is trained on ambiguous data for a particular linguistic rule. Then, we test whether the model learned a cognitively plausible rule using data disambiguating the model's generalization.
      Through this experimental scheme, we adjust whether/how the visual information helps the model infer the proper linguistic generalization.
      }
      \setlength{\belowcaptionskip}{-100pt}
      \label{fig:figure1}
\end{figure}

One general criticism of neural LMs is their lack of grounding~\cite{Roy2005-wo,Barsalou2008-fl}: they learn language solely based on text and do not model the explicit association between linguistic expressions and the associated objects/events in the real world.
This naturally leads to the hypothesis that the human--LM data efficiency gap comes from this disconnect.

In this study, we investigate whether visual information, as a representative modality promoting grounding, can accelerate the emergence of the syntactic hierarchical generalization ability of LMs, which underlies human language acquisition~\cite{chomsky1964aspects}.
Our experiments extend the single modality version of the \textit{poverty of stimulus} (POS) setting~\cite{Wilson2006-qo,Perfors2011-mu,mccoy2018revisiting,McCoy2020-mc,Warstadt2020-tk,Yedetore2023-sw} into the vision-and-language domain.
That is, we train LMs on \textit{ambiguous} image--text pairs in terms of particular linguistic rules (e.g., \textsc{Hierarchical} vs.\ \textsc{Linear} English subject--verb number agreement rules; see Figure~\ref{fig:figure1}). 
Then, we investigate whether visual input efficiently guides the models to make cognitively plausible (hierarchical) generalizations given ambiguous data, compared to text-only models.

To adjust the visual conditions, we base our experiments on either (i) realistic image--caption data~\cite{Sharma2018-wl}, or (ii) simplified, artificial data, which is a proxy for externally-guided attentional focus.
Notably, it has been argued that either strong inductive bias or additional signals, such as mutual gaze, pointing, or other forms of attentional focus, are needed to make use of multimodal input for linguistic generalization~\cite{Qu2008-fx,Johnson2012-ba} since merely adding an input modality may incur many superficial correlations and complicate rather than simplify the task~\cite{Gleitman1992-jo,Dupoux2016-ia}.
Thus, our investigation using the two types of multimodal data can be seen as an evaluation of the inductive bias of neural LMs toward multimodal linguistic generalization with and without such additional signals. Most work on grounded and situated multimodal LM as well as human language acquisition has focused on word learning~\cite{Hill2020-ye,Ma2023-tv}. In this work, we extend these investigations to the acquisition of syntactic hierarchical generalizations, the central topic toward the POS setting in NLP~\cite{mccoy2018revisiting,McCoy2020-mc}, with multimodal LMs.

In a realistic setting, we found that overall: (i) vision data \textbf{does not substantially} accelerate hierarchical generalization; (ii) this trend is consistent among 20 model settings; and (iii) this is also consistent across four different degrees of ambiguity.
In contrast, with simplified, artificial data, where visual/linguistic concepts are already abstracted and simplified, we generally found the opposite trend: vision data \textbf{did}  boost hierarchical linguistic generalization.
These contrasts suggest that neural models have the potential to make use of visual input for linguistic generalization when the visual input is made salient either through inductive bias or external signals. However, efficient generalization via more complex and ambiguous visual input is not possible in the model variants tested either because the visual processing module lacks appropriate inductive bias or the external signals of attentional salience are absent.

\section{Background}
\label{sec:background}

\subsection{Inductive bias in language acquisition}
In general, a unique generalization or rule cannot be determined solely based on the observation of finite data. The choice depends on the inductive biases of the model, such as a learner's prior knowledge~\cite{Mitchell1980-nj}.

\paragraph{In humans:}
In the context of language acquisition, it has long been argued that human learners possess a strong inductive bias due to rapid language acquisition from limited language exposure~\cite{Chomsky1980-vx,mccoy2018revisiting}.
The main question is what type of biases humans have and where these biases originate.
Regarding the former question, it has been reported that humans have a bias to prefer hierarchical generalization over linear generalization in situations like those depicted in \Cref{fig:figure1}~\cite{Crain1987-ji,Legate2002-nn}.
As for the latter question, there are two primary potential sources of inductive biases: innate factors and environmental/empirical factors.
To address this question, this study investigates the influence of a specific environmental factor --- access to visual information during language acquisition --- through computer simulations.

\paragraph{In neural models:}
Neural models typically exhibit non-human-like generalizations, such as the use of superficial cues and linear rules, as widely observed across various NLP domains~\cite{McCoy2019-ek,Warstadt2020-tk,Warstadt2020-wo,McCoy2020-mc}.
Large amounts of data are required to overcome such cognitively implausible biases during training~\cite{Warstadt2020-tk, Warstadt2020-wo}. In this context, addressing the inadequate biases in neural models and tackling their data-inefficiency issues are two aspects of the same problem. Our interest lies in understanding whether and how visual information contributes to the development of appropriate inductive bias in neural language learners.

\subsection{Hypotheses on the advantage of vision}
\label{subsec:hypothesis}
There has already been some investigation into the contribution of vision in language learning. It is important to note that this study does not take a strong position on the benefits of vision but rather conducts an exploratory investigation.

\begin{figure}[t]
    \centering
      \includegraphics[width=\linewidth]{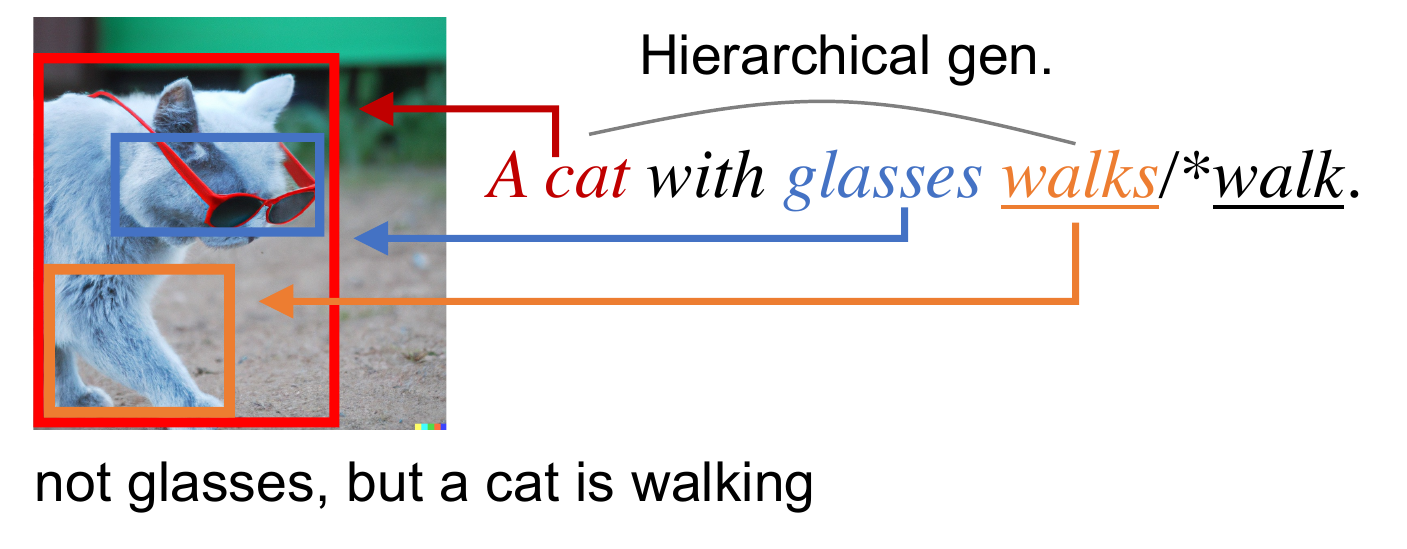}
      \caption{Images can explicate the subject--verb dependency. If a learner can ground \textit{cat}, \textit{glasses}, and \textit{walk} to their visual components, they can disambiguate that what is walking is not \textit{glasses} but \textit{cat}; such information will potentially bias the learner's language acquisition in favor of the linguistically correct rule.
      }
      \label{fig:hypothesis}
\end{figure}

\paragraph{Positive view:}
The general advantages of input beyond text modality in language acquisition have been historically emphasized~\cite{Goldberg2005-jc,Bender2020-ds}.
From an NLP perspective, the advantage of visual information typically for syntactic parsing was demonstrated~\cite{Shi2019-ot, Kojima2020-ma}.
Note that such NLP research used a specially-designed parser that already has a strong inductive bias (e.g., the training objective is parsing); our question is whether even vanilla neural models, a domain-general learner, with next-word prediction can take advantage of visual information for syntactic hierarchical generalization.
Moreover, in achieving hierarchical generalizations in settings like that illustrated in \Cref{fig:figure1}, intuitively, images have the potential to boost correct generalization.
For example, in a sentence such as \textit{a cat with glasses walks}, the information that it is the \textit{cat}, not the \textit{glasses} that is walking, could potentially bias the learning towards a hierarchical generalization.
Such a clue --- it is the \textit{cat} walking and not the \textit{glasses} --- would be explicit in the image (Figure~\ref{fig:hypothesis}) if the learner or model understands the visual concepts of \textit{cat}, \textit{glasses}, \textit{walk}, and their composition (e.g., \textit{walking cat}).
In addition, at least for the number agreement problem, the number information is, more or less, salient in the vision domain.
When the number of visual objects corresponding to grammatical subjects changes, the content of the image will change drastically, while in the text domain, only a few characters/tokens are changed.\footnote{Strictly speaking, grammatical and physical (visual) numbers are not exactly the same concepts~\cite{Spector2007-gb,Zweig2009-ff}.}

\paragraph{Negative view:}
There is also skepticism that merely providing visual input without appropriate linguistic knowledge or attentional focus could over-complicate the problem, e.g., increase the potential for superficial correlations~\cite{Gleitman1992-jo,Dupoux2016-ia}.
For example, \citet{Gleitman1992-jo} and~\citet{McDonough2011-oh} assumed that children use syntactic category information to ground words to visual input; this implies that syntactic knowledge comes first, followed by grounding.
These studies generally claim that the advantage of input beyond text in language acquisition could be driven by both humans' prior knowledge and visual input.
In this sense, if neural LMs, which are assumed to have no innate knowledge, fail to accelerate linguistic generalization with visual input, this implicitly highlights the necessity of specific learners' inductive biases or additional attentional signals in multimodal language acquisition.
Beyond syntactic generalization, there are actually some reports that visual input does not enhance the fundamental linguistic knowledge of models~\cite{Yun2021-wi,Wang2023-pb} or classifiers~\cite{Ma2021-dk}  (c.f.\ contemporaneous work by~\citet{Zhuang2024-fy} arguing multimodal input does accelerate neural LM word learning on some smaller datasets).

\paragraph{Similar attempts:}
Concurrent works have empirically investigated what linguistic ability particular neural networks can acquire solely from developmentally-plausible multimodal data that is recorded by a head-mounted camera of English-speaking children~\cite{Vong2024-ro,Qin2024-ng,Wang2023-pb}, motivated by the general, historical debates on the empiricism toward language acquisition~\cite{Elman1990-on,kirov-cotterell-2018-recurrent}.
Although their results suggest the learnability of certain linguistic properties by image-captioning models and these data, the exact advantage of visual input itself was nuanced on BLiMP~\cite{Wang2023-pb}, beyond the focus~\cite{Qin2024-ng}, or unclear~\cite{Vong2024-ro} since the evaluation tasks are image-classification/mapping, where it is somewhat obvious to see the advantage of visual input.
Furthermore, these studies examined a very limited variant of visual encoders; thus, the generality of the results was unclear.
Our evaluation potentially achieves fairer comparisons since the task itself (acceptability judgment toward syntactic generalization) is agnostic to the existence of visual modality, and we observe generally consistent results from  12 variants of vision-language models.

\section{Problem definition}
\label{sec:logic}
We briefly introduce the poverty of stimulus (POS) settings~\cite{Wilson2006-qo,Perfors2011-mu,mccoy2018revisiting,McCoy2020-mc,Warstadt2020-wo,Warstadt2020-tk,Warstadt2022-zi,Yedetore2023-sw}.
Through our experiments, we aim to quantify whether vision accelerates cognitively-plausible generalization in neural LMs.

\input{tables/examples}

\subsection{\textsc{Hierarchical} vs. \textsc{Linear} generalizations}
\label{subsec:problem}
We use the subject--verb number agreement rule as a target phenomenon.
In English, the subject and corresponding verb should match in terms of their grammatical number:

\eenumsentence{
\item \textbf{Girls} with a hat \textbf{walk}. 
\item  A \textbf{girl} with a hat \textbf{walks}. 
}

Here, Example (1b) is \textit{ambiguous} because a learner can infer at least two different generalizations from this example alone, i.e., \textsc{Hierarchical} and \textsc{Linear} rules:
\vspace{0.5cm}

\begin{dependency}[theme = simple, arc angle=15]
  \begin{deptext}[column sep=0.3em]
  (1b)  \& A girl \& with \& a \& hat \& \textbf{walks} \\
\end{deptext}
\depedge[-,label style={font=\large}]{2}{6}{\textsc{Hierarchical}}
\depedge[-,label style={font=\large,below},edge below]{5}{6}{\textsc{Linear}}
\end{dependency}

\noindent
The \textsc{Hierarchical} rule associates the grammatical number of a verb with that of its grammatical subject, while the linear one associates the number between a verb and its closest noun in a linear word order.
By contrast, Example (1a) is not ambiguous in terms of the \textsc{Hierarchical} and \textsc{Linear} rules since the number does not match under the \textsc{Linear} assumption:

\begin{dependency}[theme = simple, arc angle=15]
  \begin{deptext}[column sep=0.3em]
  (1a)  \& Girls \& with \& a \& hat \& \textbf{walk} \\
\end{deptext} 
\depedge[-,label style={font=\large}]{2}{6}{\textsc{Hierarchical}}
\depedge[-,label style={font=\large,below},edge below]{5}{6}{\textsc{*Linear} (explicit violation of number agreement)}
\end{dependency}

\vspace{0.3cm}

Our interest lies in which rule a particular learner acquires from \textit{ambiguous} data and what factors (e.g., vision) can guide the learner to prefer the \textsc{Hierarchical} rule that is linguistically correct (Section~\ref{subsec:poverty}).
The motivation for this experimental setting is further described in Section~\ref{subsec:poverty}.

We only employed this subject--verb number agreement setting in our experiments, although other studies have focused on different syntactic transformation tasks, such as question formulation or passivization~\cite{McCoy2020-mc,Warstadt2020-tk,Mueller2022-gt}.
Our motivation is the ease of collecting natural images for sentences with subject--verb agreement and the strong correlations between image entities and grammatical number. Such correlations are either absent or weak in the case of interrogative vs.\ declarative sentences and passive vs.\ active mood. 

\subsection{Poverty of stimulus setting}
\label{subsec:poverty}
Children acquire \textsc{Hierarchical} rules despite the scarcity of disambiguating sentences, like Example (1a), in real language exposure~\cite{Crain1987-ji,Legate2002-nn}.
Building on this scenario, we expose a model to (nearly) ambiguous data where the generalization cannot be determined as to whether \textsc{Linear} or \textsc{Hierarchical} rules are correct.
Then, we evaluate the model in terms of which rule is obtained from the ambiguous data via a test using unambiguous data.

\noindent
\paragraph{Data splitting strategy: }
We split data into two groups: (i) those that do not disambiguate \textsc{Linear} and \textsc{Hierarchical} rules (\textsc{Ambiguous}); and (ii) those that support the \textsc{Hierarchical} rule (\textsc{Unambiguous}).
Examples are shown in Table~\ref{tab:examples}.
Basically, the \textsc{Ambiguous} instances are used in training, and \textsc{Unambiguous} instances are used in evaluation.
We insert a few held-out \textsc{Unambiguous} instances into training data since it is counter-intuitive that a learner \textit{never} encounters direct evidence for hierarchical generalizations, i.e., \textsc{Unambiguous} instances, during language acquisition.
Therefore, we controlled the injection rate --- the extent to which disambiguating data appear during training --- for experiments analyzing sensitivity to the scarcity of direct evidence (Section~\ref{subsec:vit-base}).

\noindent
\paragraph{Model comparison:}
In this series of experiments, we compare neural models that can access visual information (\sight) and ones that do not (\blind) to assess the contribution of vision modality.
Note that ``visual information'' in this study denotes an image compatible with the meaning of a sentence, i.e., we use image--caption pairs.
The source of image caption data is described in Section~\ref{subsec:data}.

\subsection{Data}
\label{subsec:data}
We introduce two complementary data types: (i) \textsc{Natural} captions; and (ii) \textsc{Artificial} captions.
The \textsc{Natural} captions are collected from an image--caption corpus, while the \textsc{Artificial} captions are automatically created by rules to simplify the task.

\noindent
\paragraph{\textsc{Natural} dataset:}
We extracted image--caption pairs from the Conceptual Captions Corpus~\cite{Sharma2018-wl}, which is a widely-used and relatively large-scale image--caption dataset.
Specifically, we first collected captions that: (i) form a complete sentence, (ii) do not have grammatical errors\footnote{We used language-tool-python 2.7.1}; and (iii) do not have collective expressions such as \textit{family} or \textit{pair of} since these are confusing in terms of grammatical number.
Then, we split the data into the \textsc{Ambiguous} and \textsc{Unambiguous} sets using a dependency parser.\footnote{We used SpaCy~\cite{spacy}.}
Note that there might be parsing errors in this process, but we later observe that the models did not prefer the \textsc{Hierarchical} rule without injection of any disambiguating examples; this suggests that such errors do not inadvertently bias the model toward the \textsc{Hierarchical} rule.
Examples are shown in the left part of Table~\ref{tab:examples}.
The training set (\textsc{Ambiguous} part) consists of 348,861 image--caption pairs, and the unambiguous test set consists of 1,253 pairs.

\paragraph{\textsc{Artificial} dataset:}
Image--caption pairs were generated by rules.
Specifically, a caption is first generated with the template of \textcolor{purple}{\texttt{NUM1 COLOR1 SHAPE1 with NUM2 COLOR2 SHAPE2 VP}}; then, the corresponding image is automatically created (the detailed process is shown in Appendix~\ref{app:data}).
Examples are shown in the right part of Table~\ref{tab:examples}.
As with the \textsc{Natural} setting, we split the data into \textsc{Ambiguous} and \textsc{Unambiguous} cases.
Then, training and test data are created with different injection rates.
The training set (\textsc{Ambiguous} part) consists of 15,000 pairs, and the test set consists of 5,000 pairs.

This setting limits the variations of linguistic/visual concepts and sentence constructions compared to the \textsc{Natural} setting, and importantly, the alignment between linguistic and visual components can easily be extracted since the image only has visual objects related to the caption (less confounding factors), and word types and visual features have a one-to-one relationship (no lexical ambiguity; see~\cref{app:data}).
Thus, we use this artificial data setting to approximate the richer environment in which learners exploit visual inductive bias, gaze recognition, pointing and other extralinguistic signals of salience and focus to interpret otherwise ambiguous linguistic input.

\begin{figure*}
\begin{subfigure}{.5\textwidth}
  \centering
  \includegraphics[width=.95\linewidth]{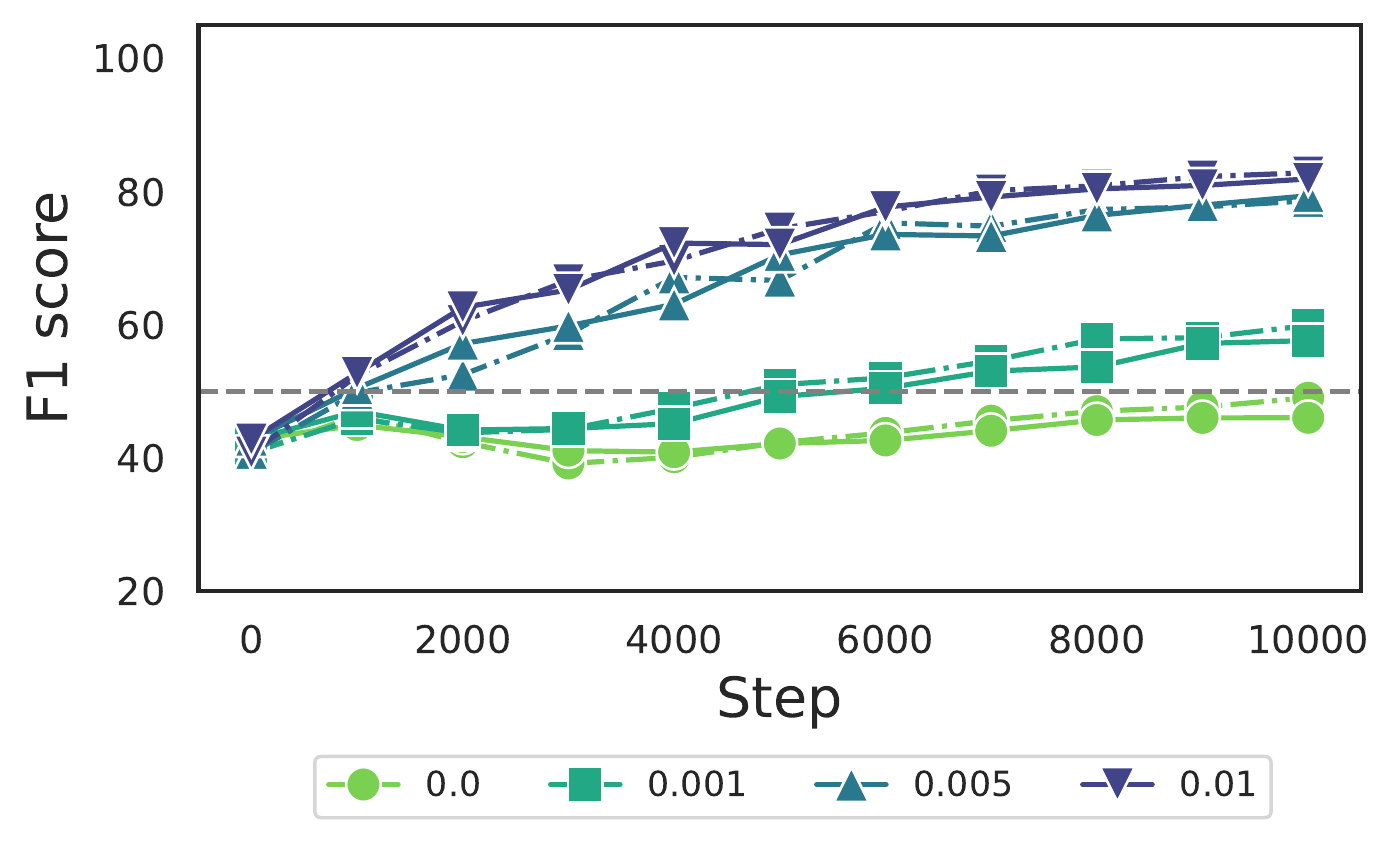}
  \caption{\textsc{Natural} setting}
  \label{fig:inoc_natural}
\end{subfigure}%
\begin{subfigure}{.5\textwidth}
  \centering
  \includegraphics[width=.95\linewidth]{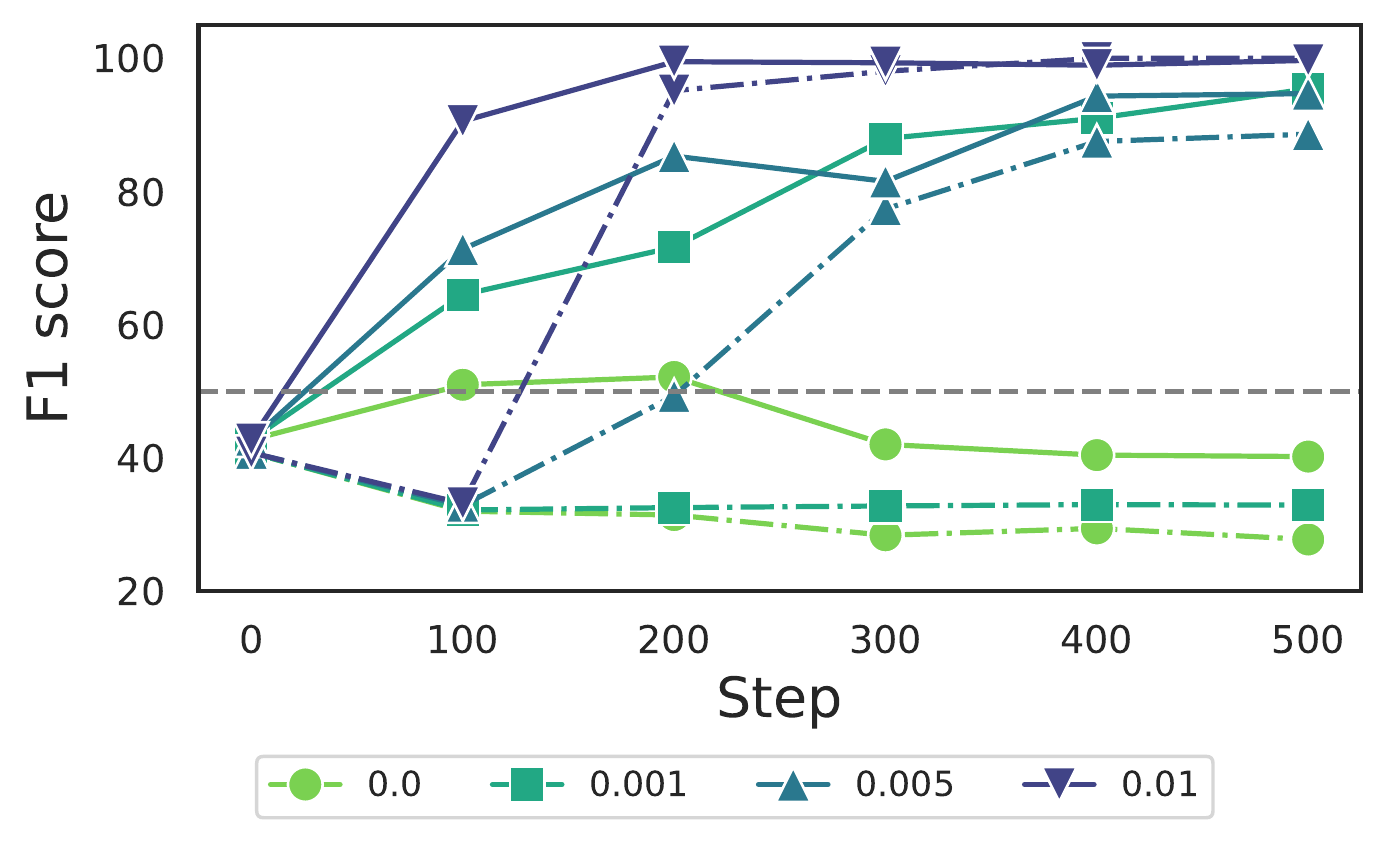}
  \caption{\textsc{Artificial} setting}
  \label{fig:inoc_art}
\end{subfigure}
\caption{Generalization performance of the model initialized with Vit-base. The $x$-axis denotes the parameter update steps, and the $y$-axis denotes the preference for the \textsc{Hierarchical} generalization rule (F1 scores multiplied by 100). We adopted four settings with different injection rates of \{0, 0.001, 0.005, 0.01\}. The normal lines correspond to the model with visual input (\sight), and the dashed lines correspond to the preference of those without visual input (\blind).  The chance rate of the F1 score is 50.}
\label{fig:inoc}
\end{figure*}

\subsection{Evaluation}
For each \textsc{Unambiguous} instance, we prepared two candidate captions differing only in the verb's grammatical number (e.g., \textit{two red rectangles with a black circle \textbf{play/plays} soccer}); one aligns with the \textsc{Hierarchical} rule, and the counterfactual one with the \textsc{Linear} rule by modifying the grammatical number of its main verb. The model's generalization preference is determined by which caption has a higher probability.

Specifically, a model $\theta$ computes the probabilities of each caption $\bm s=[w_1, \cdots, w_n]$ given the corresponding image $v$:
\begin{equation}
p(\bm s|v) = \prod_{t=1}^{n} p_{\theta}(w_t | \bm w_{<t}, v) \;\;,
\end{equation}
where $\bm w_{<t}$ denotes the left context of $w_t$ in the caption $\bm s$. We calculated the macro-F1 score, considering the inflection corresponding to the \textsc{Hierarchical} rule as correct and treating the task as a binary classification problem for selecting a grammatically-correct sentence.
As we are interested in language acquisition efficiency, we report F1 scores at various intermediate training steps.
 
\subsection{Models}
We use the Transformer seq2seq image-caption model as a vision-and-language model \sighted, with the encoder set as a pre-trained vision encoder like ViT~\cite{Dosovitskiy2020-cu}. 
An image is input to the encoder, and the decoder predicts the caption in a left-to-right manner, accessing visual information via cross-attention. 
Intuitively, this can be viewed as a sentence-level LM that can access visual information. 
For the image-less \blind model, we replaced the input image with a white noise image during training and inference. Models are trained with cross-entropy loss to generate the reference caption.
The vision encoder is further updated during the training.

We adopted the GPT-2 small (124M) architecture~\cite{Radford_undated-nn} for the decoder, with parameters randomly initialized, considering a language acquisition scenario from scratch. As an encoder, we initially used Vit-base~\cite{Dosovitskiy2020-cu} in Section~\ref{subsec:vit-base} and further examined various encoders in Section~\ref{subsec:encoders} to enhance the generality of the conclusion. Hyperparameters are listed in Appendix~\ref{app:encoders}. In each setting, we train two models with different seeds and report the average score.

\section{Experiments}

\subsection{Generalization preferences}
\label{subsec:vit-base}
We first analyze the model using the pre-trained Vit-base encoder.
We examined four different injection rates of \{0, 0.001, 0.005, 0.01\}; for example, the rate 0.001 means that ten held-out \textsc{Unambiguous} instances are added into the training data if the original training data size is 10,000.

\noindent
\paragraph{Results:}
The results are shown in Figure~\ref{fig:inoc}, with scores averaged across models with different seeds. 
These indicate the following:
\begin{itemize}
\setlength{\parskip}{0cm}
\setlength{\itemsep}{0.1cm}
\item In the \textsc{Natural} setting, visual inputs do not generate a substantial difference in generalization efficiency.
\item In the \textsc{Artificial} setting, visual inputs accelerate hierarchical generalization, especially at the early stages of learning. 
\item At the initial stage of learning in the \textsc{Natural} and \textsc{Artificial} settings with a low injection rate, the \textsc{Linear} rule emerged (F1-score below chance rate), indicating that the model originally has a \textsc{Linear} bias. This is consistent with existing studies in the text-only domain~\cite{McCoy2020-mc}.
\item With moderate rates of injection, e.g., above the rate of 0.005, the models gradually acquired the \textsc{Hierarchical} rule, showing sensitivity to the slight bias in data distribution.
\end{itemize}

We further discuss the implications of the contrasting results between the \textsc{Natural} and \textsc{Artificial} settings in Section~\ref{sec:discussion}.

\input{tables/encoders}

\begin{figure}[t]
    \begin{minipage}[t]{\hsize}
    \centering
      \includegraphics[width=\linewidth]{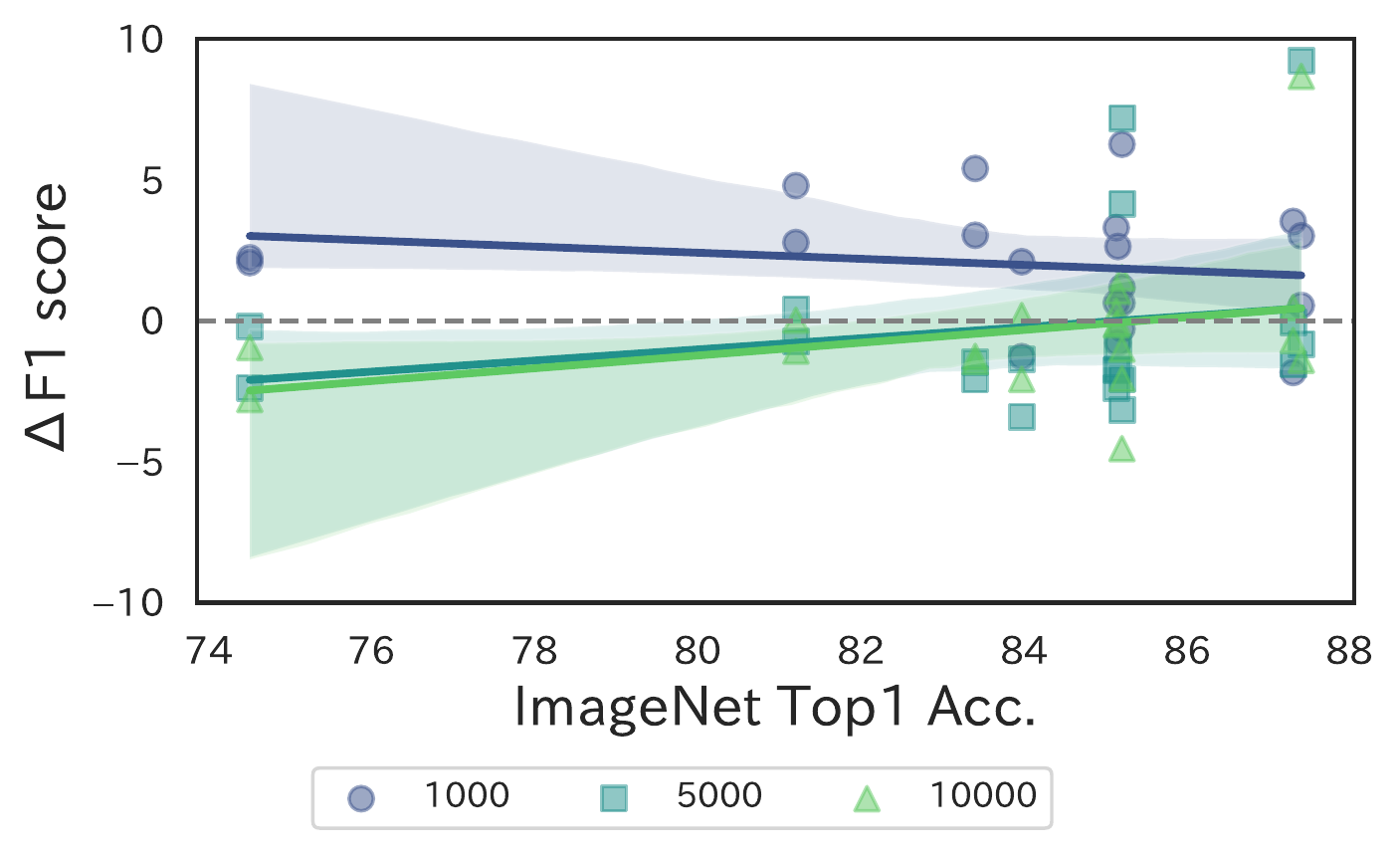}
      \subcaption{Relationship between encoders' ImageNet accuracy (x-axis) and their advantage in \textsc{Hierarchical} generalization (F1 score difference of \sight\ $-$ \blind; y-axis). The F1 score is measured at several checkpoints during training (1000, 5000, and 10000).
    }
    \label{fig:imagenet}
    \end{minipage} \\
    \begin{minipage}[t]{\hsize}
    \centering
      \includegraphics[width=\linewidth]{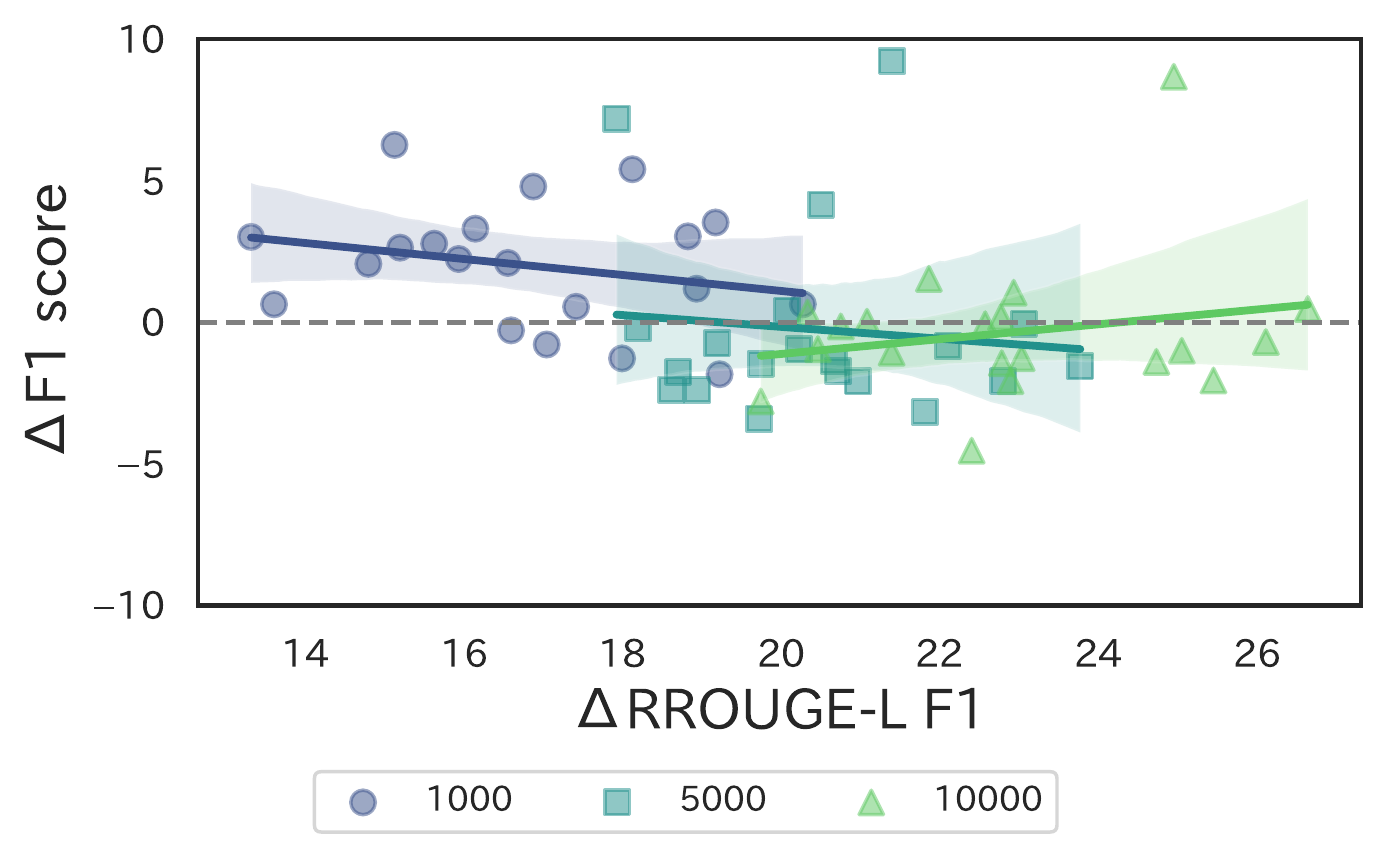}
      \subcaption{Relationship between encoders' captioning performance in the validation set (x-axis) and their advantage in \textsc{Hierarchical} generalization (F1 score difference of \sight\ $-$ \blind; y-axis). These scores are measured at several checkpoints during training (1000, 5000, and 10000).
    }
    \label{fig:rouge}
    \end{minipage} \\
    \caption{Relationship between CV-oriented metrics and the contribution to \textsc{Hierarchical} generalization in the \textsc{Natural} setting. Each dot corresponds to each setting \{10 encoders\}$\times$\{2 seeds\}$\times$\{3 training steps\}, and its color/shape corresponds to training steps.
    }
     \label{fig:cv_metrics}
\end{figure}

\subsection{Vision encoder variations}
\label{subsec:encoders}
To investigate whether our results are specific to a particular model setting, we further analyze ten vision-language models with different encoder-decoder settings, demonstrating general consistency across various settings.

\paragraph{Generality of the (in)effectiveness of vision:}
We tested the models using ten different vision encoders: Vit-\{base, large, xlarge\}~\cite{Dosovitskiy2020-cu}, Beit-\{base, large\}~\cite{Bao2022-ha}, Deit-\{base, small, tiny\}~\cite{Touvron2021-at}, and Swin-\{base, large\}~\cite{Liu2021-qc}.
We also examined two baselines: one using randomly initialized Vit-base (Scratch) and a model using the pre-trained GPT-2~\cite{Radford_undated-nn} as a decoder (Vit-GPT2).
Note that the Vit-GPT2 model is already trained on large-scale text data, including disambiguating sentences; thus, it is not surprising that they achieve hierarchical generalization.
We fix the inoculation rate to 0.01 in this Section.

The results are summarized in Table~\ref{tbl:results}.
The observations are similar to those in Section~\ref{subsec:vit-base}: (i) the effect size of the visual input factor is larger in the \textsc{Artificial} setting than the \textsc{Natural} setting, especially at the early stage of learning;\footnote{With a two-sided Wilcoxon rank-sum test, the $\Delta$ scores from the 100-step \textsc{Artificial} setting was significantly larger than those in the 1000-step setting across models and seeds ($p=5.3\mathrm{e}{-4}<0.05$).} (ii) vision data generally has a positive/negative effect on the generalization at the early/late stage.\footnote{With a two-sided one-sample t-test, the $\Delta$ scores were significantly larger than zero across models and seeds in the 1,000-step \textsc{Natural} setting ($p=4.1\mathrm{e}{-4}<0.05$) and 100-step \textsc{Artificial} setting ($p=1.8\mathrm{e}{-5}<0.05$), not significant in the 5,000/10,000-step \textsc{Natural} settings ($p=0.6$, $p=0.4$), and lower than zero in the 500-step \textsc{Artificial} setting ($p=8.0\mathrm{e}{-3}<0.05$).} 
Note that models with visual input (\sight) achieved ROUGE-L F1 scores of $30$--$40$ in the \textsc{Natural} setting (Appendix~\ref{app:encoders}), whereas those without visual input (\blind) yielded the scores of around $15$; this improvement indicates that the models do not ignore visual input.

As minor points, Beit-based models yielded somewhat idiosyncratic trends (\textsc{Hierarchical} generalization is hurt at the late stage in the \textsc{Artificial} setting).
In addition, as a sanity check, Vit-GPT2, which is pre-trained over a massive amount of text data, achieved almost perfect hierarchical generalization from the early stages of training in both \textsc{Natural} and \textsc{Artificial} settings.

\paragraph{Which vision encoder relatively accelerates hierarchical generalization?}
Different vision encoders generally show a similar trend, but the degree of their advantage is slightly different---what kind of encoder benefits most from vision inputs?
This can be viewed as an evaluation of vision encoders from a cognitive perspective.
Figure~\ref{fig:cv_metrics} shows the following: (i) no clear relationship between the encoders' ImageNet top-1 accuracy\footnote{We used the scores reported in their original papers.} and their contribution to linguistic \textsc{Hierarchical} generalization (\diff\ F1 score in Table~\ref{tbl:results}); and (ii) no clear relationship between image--captioning performance and the contribution to hierarchical generalization.
Note that the $\Delta$ROUGE in Figure~\ref{fig:rouge} indicates the ROUGE gain from a model without visual input to the one with visual input based on the same architecture.
The results indicate that an engineeringly better vision encoder does not always lead to better linguistic generalization when combined with a language decoder.

\input{tables/difficult}

\section{Discussion and limitations}
\label{sec:discussion}

\noindent
\paragraph{Mixed results in \textsc{Natural} and \textsc{Artificial} settings:}

The limited advantage of vision in the \textsc{Natural} setting suggests at least two possibilities: (i) vision is not helpful for efficient language acquisition; or (ii) vision is potentially helpful in human language acquisition scenario, but neural models lack certain human-like biases, such as learners' prior knowledge or training/data scenario related to vision-language grounding.
If one accepts the general argument about the advantage of vision and/or the advantage in the \textsc{Artificial} setting as a support for the potential usefulness of visual input, vision is useful in linguistic generalization --- and interpretation (ii) is plausible.
Thus, the challenge lies in how the learner can extract meaningful intake from raw images and texts, and at least the modern neural models we examined might not possess such an ability. 
This view aligns with the considerations put forth by, for example, \citet{Gleitman1992-jo} and \citet{Dupoux2016-ia}.

\noindent
\paragraph{Words beyond the image content:}
What specific difficulties exist in the \textsc{Natural} data?
One potential challenge we considered based on the dataset is that the natural caption contains information that is \textbf{not} present in the image, which might cause confusion in terms of the visual grounding of the sentence. For example, the first image in Table~\ref{tab:difficult} has a caption \textit{the walls over the toilet need a small cabinet}. In this case, the \textit{cabinet} is not in the image, although it is not directly relevant to the subject--verb agreement. The second example's caption in Table~\ref{tab:difficult} also mentions objects beyond the image; here, the word \textit{boys} does not refer to the boy in this image but any boy with similar eyes to him. This is potentially confusing in terms of number agreement since the grammatical subject is in plural form, but the image shows one object.
These assert that visual grounding already needs linguistic knowledge and the question of where such linguistic knowledge should come from.

\noindent
\paragraph{Coverage of the experiments:}
We only focused on a specific syntactic phenomenon, subject--verb number agreement rule.
Extending the experimental settings to cover broader linguistic phenomena, e.g., including revisiting vocabulary acquisition~\cite{rasanen2019computational}, is needed to draw more general conclusions.
In Appendix~\ref{app:blimp}, we conducted a preliminary examination using the BLiMP benchmark~\cite{Warstadt2020-or} on the linguistic knowledge of models with/without vision; this also implied that visual input alone does not lead to a substantial advantage.
Nevertheless, typical resources for linguistic probes, including BLiMP, use only text input; it is not obvious how to use such data to evaluate multimodal models.
We hope that this study encourages the community to build a dataset to probe the fine-grained linguistic knowledge of multimodal models.

\section{Conclusions}

We conducted two complementary experiments --- a noisy, realistic setting and a simplified, artificial one --- to investigate the advantage of vision in the syntactic generalization of LMs.
Our results showed that vision accelerates proper linguistic generalization under a simplified setting, but LMs struggled with proper generalization based on noisy, realistic data.
These mixed results suggest several possibilities; for example, an image can potentially boost language acquisition, but neural learners may require additional visual/linguistic prior knowledge or externally-provided attentional focus to robustly make use of \textit{raw} images for efficient language acquisition.

\section*{Limitations}
In addition to the limitations of our work raised in~\cref{sec:discussion}, the following are potential concerns.
First, the data size is relatively small; the training data in the \textsc{Natural} setting consists of around 3.5M tokens.
Nevertheless, experiments with similar motivations have been conducted with the same or smaller scale of dataset~\cite{Nikolaus2019-ay,Wang2023-pb}.
Furthermore, at least based on the report that human infants around 18 months learn syntactic dependencies~\cite{Perkins2021-jy} and they are typically exposed to 2--7M words per year~\cite{Gilkerson2017-iz}, our data size may not be too small to learn syntactic rules.

Second, we only focused on a specific type of vision-language model---image-captioning models.
There are other formulations involving vision-and-language interaction, such as text-to-image models~\cite{Ramesh2021-sy}, discrimination models like CLIP~\cite{Radford2021-bt}, or more generally, LMs with a visual input support~\cite{Alayrac2022-vo,achiam2023gpt}.
Investigating the inductive bias related to such architectural/task differences would be an interesting direction for future work.
Evaluating larger models will also provide us with insights into scaling laws in this context.
Having said that, such experiments require more computing resources than a typical laboratory has, which was an unrealistic direction for us to explore.
More generally, humans see both static and dynamic input during language acquisition.
Therefore, extension from image to video is an important future direction of research.

Third, there are concurrent endeavors to examine the contribution of visual information to proper linguistic generalizations of neural LMs from cognitively-motivated perspectives~\cite{Wang2023-pb,Zhuang2024-fy}; the closest initiative would be the 2nd-round of the BabyLM shared task, which includes multimodal data~\cite{Choshen2024-sn}.
Enhancing the connection to such recent works will be the target of future work, and we would like to highlight that our study has employed a control to the training data properties to gain rich insights into the model's inductive biases, which has rarely been achieved in existing multimodal experiments and is orthogonal to the holistic evaluation of pretrained vision-language models.

\section*{Ethical concerns}
This study employed a widely-used, publicly available image--caption dataset, to avoid ethical concerns.
In our argument, we assumed that humans usually have access to visual information during language acquisition; this is not intended to discriminate against vision-impaired people. Our general interest is in grounding, which can also be established by other modalities, and we focus on the vision modality as one case study.
Perhaps our results of no advantage of visual input may be supported by the success of human language acquisition regardless of their congenital blindness; such a broader connection to human language acquisition should be enhanced in future work.

\section*{Acknowledgement}
This work was partially supported by JST CREST Grant Number JPMJCR20D2, Japan.
We sincerely appreciate anonymous reviewers, including those for our previous versions, for their knowledgeful comments.
We appreciate Ted Briscoe and Yova Kementchedjhieva for their insightful feedback on the early version of this paper.
We also thank the Tohoku NLP Group members, especially Kentaro Inui, for their constructive comments on our earlier work.

\bibliography{custom}

\clearpage

\appendix
\section*{Appendix}

\section{Artificial data}
\label{app:data}
Table~\ref{tbl:data} shows the textual and visual features used in the \textsc{Artificial} dataset.
The \textcolor{purple}{\texttt{NUM2 COLOR2 SHAPE2}} objects are placed on top of each \textcolor{purple}{\texttt{NUM1 COLOR1 SHAPE1}} object, and the \textcolor{purple}{\texttt{VP}} object is overlaid on the \textcolor{purple}{\texttt{NUM1 COLOR1 SHAPE1}} object.
We created $3\times3\times5\times4\times4\times4\times10$=28,800 image--caption pairs; 15,000 data are used for training, 1,000 data are used for validation, and 5,000 data are used for evaluation (we sampled 21,000 instances from the 28,800 data).

\input{tables/artificial.tex}

\section{Vision encoders}
\label{app:encoders}
All the encoders we used are available in Huggingface.
These are pre-trained/fine-tuned on the ImageNet-21k(22k) data with $224^2$ resolution and batch size of $16$.
Table~\ref{tbl:hyperparam} shows the common hyperparameters across the models; other encoder hyperparameters follow the original pre-trained model.
To avoid over-fitting, we applied RandAugemnt~\cite{Cubuk2020-nt} to the input image and replaced the input image with a white noise with a probability of 0.2.
Table~\ref{tbl:rouge} shows the image--captioning performance of each model in the validation split of \textsc{natural} data.\footnote{Hold-out 1000 \textsc{Ambiguous} instances that do not overlap with the training data.}
The ROUGE score is computed using the implementation of \url{https://huggingface.co/spaces/evaluate-metric/rouge}.
The exact pre-trained models we used are as follows:
\vspace{0.2cm}

\input{tables/blimp.tex}

\input{tables/parameters.tex}

\input{tables/rouge.tex}

\noindent
\textbf{Vit:}
\begin{itemize}
    \item \url{https://huggingface.co/google/vit-base-patch16-224-in21k}
    \item \url{https://huggingface.co/google/vit-large-patch16-224-in21k}
    \item \url{https://huggingface.co/google/vit-huge-patch14-224-in21k}
\end{itemize}

\noindent
\textbf{Beit:}
\begin{itemize}
\item \url{https://huggingface.co/microsoft/beit-base-patch16-224-pt22k-ft22k} \item \url{https://huggingface.co/microsoft/beit-large-patch16-224-pt22k-ft22k}
\end{itemize}

\textbf{Deit:}
\begin{itemize}
\item \url{https://huggingface.co/facebook/deit-base-distilled-patch16-224}
\item \url{https://huggingface.co/facebook/deit-small-distilled-patch16-224}
\item \url{https://huggingface.co/facebook/deit-tiny-distilled-patch16-224}
\end{itemize}

\noindent
\textbf{Swin:}
\begin{itemize}
\item \url{https://huggingface.co/microsoft/swin-base-patch4-window7-224-in22k}
\item \url{https://huggingface.co/microsoft/swin-large-patch4-window12-384-in22k}
\end{itemize}

\section{Evaluation on BLiMP benchmark}
\label{app:blimp}
We evaluate linguistic knowledge in models with/without vision using the BLiMP benchmark, which has several ``circuits'' targeting specific linguistic knowledge.
Each instance in the circuit is a minimally different sentence pair regarding the targeted grammar item.
Similar to our experiment, we observed whether a model could assign a lower perplexity\footnote{Sentence pairs in the BLiMP sometimes have different lengths; thus, we avoid using a vanilla probability.} to the grammatically correct sentence.

BLiMP has only text input; thus, we must input a sentence alone (and a white noise image) to vision-language models.
When inputting only text, a model without vision \blind\ might be unfairly favored over a model with vision \sight\ from the perspective of the training--inference gap.
 To achieve a fairer comparison, we also introduce another baseline without proper visual grounding \cyclone\ that is trained with \textit{randomly shuffled} image--caption pairs.
 We intend that \sight\ and \cyclone\ models suffer from a similar degree of handicap regarding the training--inference gap.
 
Table~\ref{tbl:blimp} shows accuracies on each circuit of BLiMP.
Vit-base encoder models were evaluated, which are trained using the training set of \textsc{Natural} data with 10,000 parameter updates.
The model with vision \sight\ does not show a substantial advantage over \blind\ and \cyclone\ baselines; this implies that visual input alone cannot enhance their linguistic knowledge.

\end{document}

%% file: tables/examples.tex
{\setlength{\fboxsep}{1pt}
\begin{table*}[ht]
\centering
\footnotesize{
\begin{tabular}{p{2cm}cp{4cm}cp{4cm}} \toprule
\diagbox[width=2.2cm]{\footnotesize Split}{\footnotesize Setting} & \multicolumn{2}{c}{\textsc{Natural}} & \multicolumn{2}{c}{\textsc{Artificial}} \\
\cmidrule(r){1-1} \cmidrule(lr){2-3} \cmidrule(l){4-5}
\multirow{4}{2cm}{\begin{tabular}{l} \textsc{Ambigu} \\ \textsc{-ous}\end{tabular}} &\includegraphics[align=c, width=0.1\linewidth]{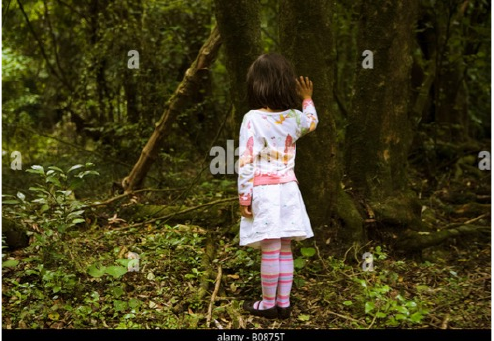} &
\begin{tabular}{l} \textit{\colorbox{mycyan}{\textcolor{orange}{girl}} aged \textcolor{red}{stands} with a} \\ \textit{hand on a tree alone} \end{tabular}
 & \includegraphics[align=c, width=0.15\linewidth]{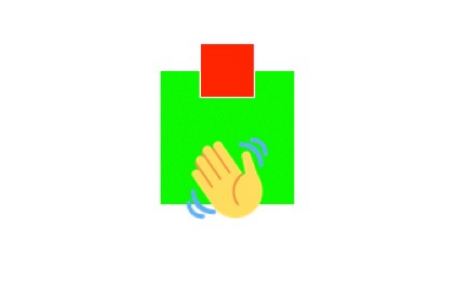} & \begin{tabular}{l} \textit{a lime \textcolor{orange}{rectangle} with a red} \\ \textit{\colorbox{mycyan}{rectangle} \textcolor{red}{waves} its hand} \end{tabular}
 \\
&\includegraphics[align=c, width=0.1\linewidth]{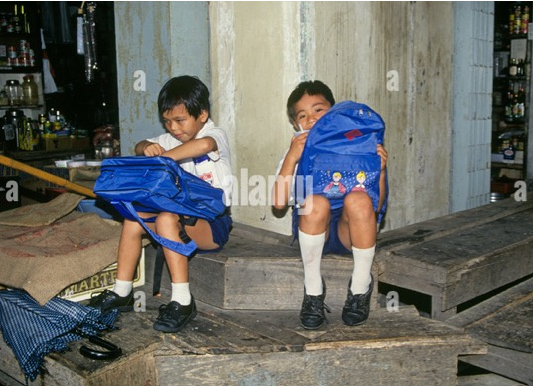} & \begin{tabular}{l} \textit{\textcolor{orange}{young boys} with school} \\ \textit{uniforms and \colorbox{mycyan}{backpacks}} \\ \textit{\textcolor{red}{prepare} for school on an} \\ \textit{early morning} \end{tabular}
 & \includegraphics[align=c, width=0.15\linewidth]{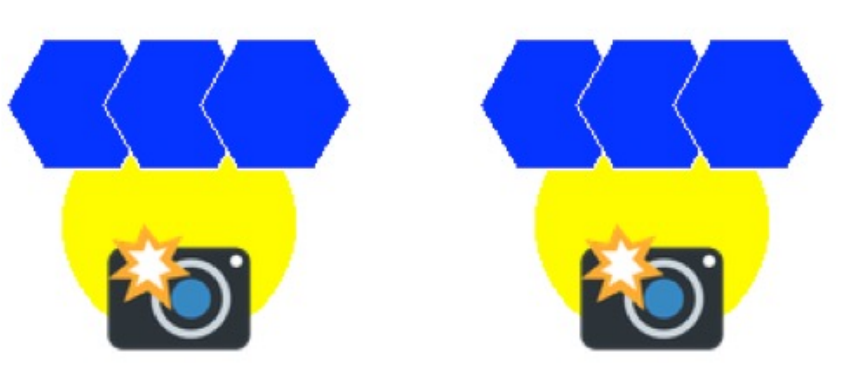} & \begin{tabular}{l} \textit{two yellow \textcolor{orange}{circles} with three} \\ \textit{blue \colorbox{mycyan}{hexagons} \textcolor{red}{take} a photo} \end{tabular} \\
 \cmidrule(r){1-1} \cmidrule(lr){2-3} \cmidrule(l){4-5}
\begin{tabular}{l} \textsc{DisAmbigu}\\-\textsc{ating}\end{tabular} &\includegraphics[align=c, width=0.1\linewidth]{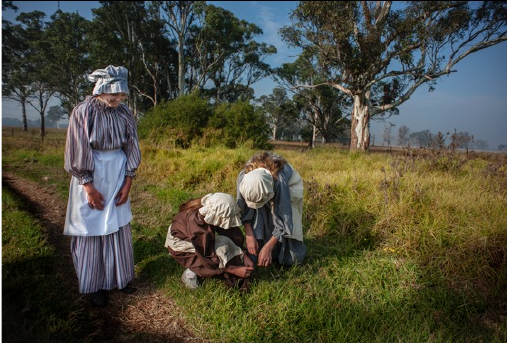} & \begin{tabular}{l} \textit{young \textcolor{orange}{girls} dressed in} 
 \\ \textit{colonial \colorbox{mycyan}{gear} \textcolor{red}{tie} their} \\ \textit{shoes at farm} \end{tabular} & \includegraphics[align=c, width=0.15\linewidth]{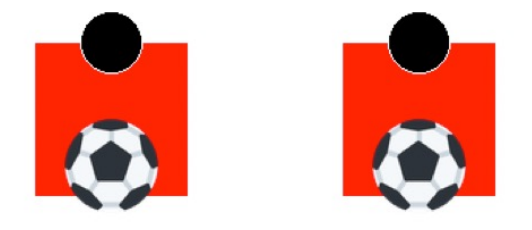} & \begin{tabular}{l} \textit{two red \textcolor{orange}{rectangles} with a black} \\ \textit{\colorbox{mycyan}{circle} \textcolor{red}{play} soccer} \end{tabular} \\
\bottomrule
\end{tabular}
}
\caption{
Examples of image-caption pairs. The \textsc{Natural} data is collected from conceptual captions corpus, and the \textsc{Artificial} data is generated by rules. In the \textsc{Ambiguous} set, the grammatical numbers of \textcolor{red}{verb}, its corresponding \textcolor{orange}{subject}, and its immediately preceding \colorbox{mycyan}{noun} are identical; in this sense, they are ambiguous toward which is the correct rule of number agreement, \textsc{Linear} or \textsc{Hierarchical}. By contrast, the \textsc{DisAmbiguating} instances disambiguate the rule.
}
\label{tab:examples}
\end{table*}
}

%% file: tables/encoders.tex
{\setlength{\fboxsep}{1pt}
\begin{table}[t]
{\footnotesize
\centering
\tabcolsep  1.5pt
\begin{center}
\begin{tabular}{p{1.3cm}crrrrr} \toprule
&& \multicolumn{3}{c}{\textsc{Natural}}  & \multicolumn{2}{c}{\textsc{Artificial}} \\
\cmidrule(lr){3-5} \cmidrule(lr){6-7} 
 Models & Vision & 1,000 & 5,000 & 10,000 & 100& 500 \\
\cmidrule(r){1-1} \cmidrule(lr){2-2} \cmidrule(lr){3-5} \cmidrule(lr){6-7}
\multirow{2}{1.5cm}{Vit-base (86M)} & \sighted & $52.8$ & $72.0$ & $81.9$ & $90.6$ & $99.7$ \\
& \diff &\gcella $+0.41$ & \rcellb $-2.38$ & \rcella $-0.94$ & \gcellc $\mathbf{+57.4}$ & \rcella $-0.31$ \\
\cmidrule(r){1-1} \cmidrule(lr){2-2} \cmidrule(lr){3-5} \cmidrule(lr){6-7}
\multirow{2}{1.5cm}{Vit-large (307M)}& \sighted & $52.9$ & $74.9$ & $83.1$ & $52.6$ & $92.2$ \\
& \diff & \gcella $+0.93$ & \rcellb $-1.13$ & \gcella $+0.65$ & \gcellc $\mathbf{+19.4}$ & \rcellb $-7.76$ \\
\cmidrule(r){1-1} \cmidrule(lr){2-2} \cmidrule(lr){3-5} \cmidrule(lr){6-7}
\multirow{2}{1.5cm}{Vit-huge (632M)}& \sighted & $52.6$ & $73.9$ & $82.6$ & $42.6$ & $100$ \\
& \diff & \gcellb $+1.98$ & \rcellb $-2.07$ & \gcella $+0.10$ & \gcellb $+9.21$ & \grcell $0.00$ \\
\cmidrule(r){1-1} \cmidrule(lr){2-2} \cmidrule(lr){3-5} \cmidrule(lr){6-7}
\multirow{2}{1.5cm}{Beit-base (86M)}& \sighted & $46.7$ & $59.0$ & $66.4$ & $45.8$ & $74.8$ \\
& \diff & \gcellb $+2.99$ & \gcellb $+5.68$ & \rcellb $-1.50$ &  \gcellc $\mathbf{+11.7}$ & \rcellc $-25.0$ \\
\cmidrule(r){1-1} \cmidrule(lr){2-2} \cmidrule(lr){3-5} \cmidrule(lr){6-7}
\multirow{2}{1.5cm}{Beit-large (307M)}& \sighted & $45.6$ & $65.3$ & $73.3$ & $38.3$ & $57.7$ \\
& \diff & \gcellb $+1.57$ & \gcellb $+4.32$ & \gcellb $+3.80$ &  \gcellb $+5.09$ & \rcellc $-38.4$ \\
\cmidrule(r){1-1} \cmidrule(lr){2-2} \cmidrule(lr){3-5} \cmidrule(lr){6-7}
\multirow{2}{1.5cm}{Deit-base (86M)}& \sighted & $54.9$ & $72.5$ & $81.2$  & $67.4$ & $99.9$ \\
& \diff & \gcellb $+4.23$ & \rcellb $-1.77$ & \rcellb $-1.35$ & \gcellc $\mathbf{+32.9}$ & \gcella $+0.08$ \\
\cmidrule(r){1-1} \cmidrule(lr){2-2} \cmidrule(lr){3-5} \cmidrule(lr){6-7}
\multirow{2}{1.5cm}{Deit-small (22M)}& \sighted & $52.9$ & $73.7$ & $83.2$  & $73.1$ & $94.1$  \\
& \diff & \gcellb $+3.79$ & \rcella $-0.16$  & \rcella $-0.52$ & \gcellc $\mathbf{+27.1}$ & \rcellb $-5.86$ \\
\cmidrule(r){1-1} \cmidrule(lr){2-2} \cmidrule(lr){3-5} \cmidrule(lr){6-7}
\multirow{2}{1.5cm}{Deit-tiny (5M)}& \sighted & $52.6$ & $73.5$ & $81.0$  & $88.8$ & $87.8$  \\
& \diff & \gcellb $+2.16$ & \rcellb $-1.29$  & \rcellb $-1.87$ & \gcellc $\mathbf{+32.5}$ & \rcellc $-12.2$ \\
\cmidrule(r){1-1} \cmidrule(lr){2-2} \cmidrule(lr){3-5} \cmidrule(lr){6-7}
\multirow{2}{1.5cm}{Swin-base (88M)}& \sighted & $53.0$ & $73.0$ & $81.8$  & $80.5$ & $100$  \\
& \diff & \gcella $+0.92$ & \rcellb $-2.61$  & \rcellb $-1.05$ & \gcellc $\mathbf{+33.2}$ & \grcell $0.00$ \\
\cmidrule(r){1-1} \cmidrule(lr){2-2} \cmidrule(lr){3-5} \cmidrule(lr){6-7}
\multirow{2}{1.5cm}{Swin-large (197M)}& \sighted & $53.3$ & $73.9$ & $82.4$  & $74.9$ & $100$  \\
& \diff & \gcella $+0.85$ & \rcella $-0.79$  & \rcella $-0.11$ & \gcellc $\mathbf{+39.3}$ & \grcell $0.00$ \\
\midrule
\multirow{2}{1.5cm}{Scratch (86M)}& \sighted & $49.3$ & $72.6$ & $81.0$  & $50.7$ & $100$ \\
& \diff & \gcellb $+1.75$ & \rcellb $-3.22$ & \rcellb $-1.62$ & \gcellb $+5.10$ & \grcell $0.00$ \\
\cmidrule(r){1-1} \cmidrule(lr){2-2} \cmidrule(lr){3-5} \cmidrule(lr){6-7}
\multirow{2}{1.5cm}{Vit-GPT2 (86M)}& \sighted & $95.6$ & $97.0$ & $96.6$ & $90.8$ & $100$ \\
& \diff & \grcell $+0.04$ & \gcellb $+0.18$ & \rcella $-0.11$ & \rcellb $-9.21$ &\grcell $0.00$ \\
\bottomrule
\end{tabular}
\end{center}
}
\caption{The preference for \textsc{Hierarchical} generalization (F1 score) of models without various configurations. F1 scores are multiplied by 100. 
The column names such as 1,000, 5,000, and 10,000 denote the training steps. 
Scores in the \sighted\ row indicate the results of models with visual inputs \sight, and those in \diff\ indicate the score difference between models with and without visual inputs (\sight $-$ \blind). 
}
\label{tbl:results}
\end{table}
}

%% file: tables/difficult.tex
{\setlength{\fboxsep}{1pt}
\begin{table}[t]
\centering
\small{
\begin{tabular}{cp{4cm}} \toprule
\includegraphics[align=c, width=0.2\linewidth]{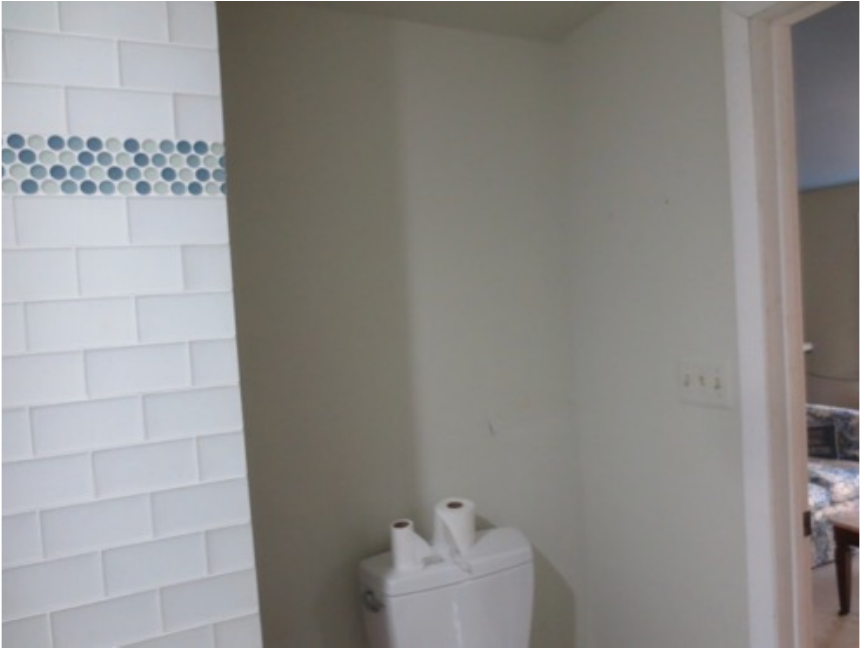} & \textit{the walls over the toilet need a small cabinet}
 \\
 \includegraphics[align=c, width=0.2\linewidth]{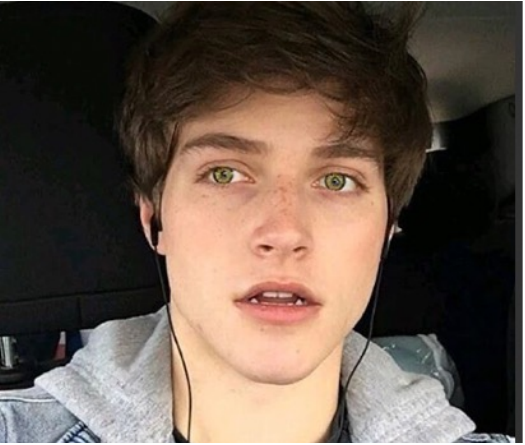} &  \textit{boys with eyes like that drive me crazy} \\ 
\bottomrule
\end{tabular}
}
\caption{Examples exhibiting some challenging features of \textsc{Natural} image captions.}
\label{tab:difficult}
\end{table}
}

%% file: tables/artificial.tex
\begin{table}[h]
{\small
\centering
\begin{tabular}{lp{3cm}c} \toprule
 Category & Word & Visual feature \\
\cmidrule(lr){1-1} \cmidrule(lr){2-2} \cmidrule(lr){3-3}
\multirow{3}{1.5cm}{\textcolor{purple}{\texttt{NUM1/2}}} & a & \colorbox{black}{ }\\
& two & \colorbox{black}{ } \colorbox{black}{ } \\
& three & \colorbox{black}{ } \colorbox{black}{ } \colorbox{black}{ } \\
\cmidrule(lr){1-1} \cmidrule(lr){2-2} \cmidrule(lr){3-3}
\multirow{5}{1.5cm}{\textcolor{purple}{\texttt{COLOR1/2}}} & black & \colorbox{black}{ } \\
& red & \colorbox{red}{ }\\
& blue & \colorbox{blue}{ } \\
& yellow & \colorbox{yellow}{ } \\
& lime & \colorbox{lime}{ } \\
\cmidrule(lr){1-1} \cmidrule(lr){2-2} \cmidrule(lr){3-3}
\multirow{4}{1.5cm}{\textcolor{purple}{\texttt{SHAPE1/2}}} & circle(s) & \includegraphics[align=c, width=1em]{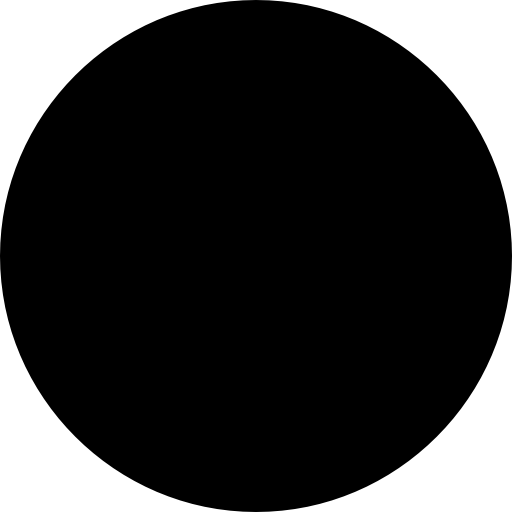} \\
& rectangle(s) &\includegraphics[align=c, width=1em]{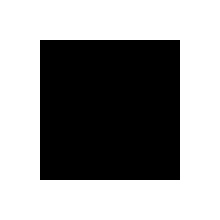} \\
& triangle(s) & \includegraphics[align=c, width=1em]{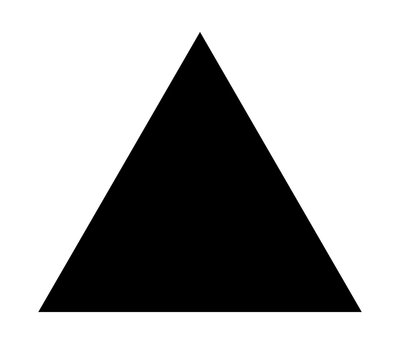} \\
& hexagon(s) & \includegraphics[align=c, width=1em]{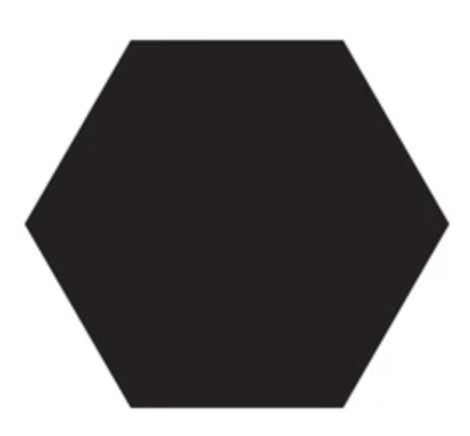} \\
\cmidrule(lr){1-1} \cmidrule(lr){2-2} \cmidrule(lr){3-3}
\multirow{10}{1.5cm}{\textcolor{purple}{\texttt{VP}}} & walk(s) & \includegraphics[align=c, width=2em]{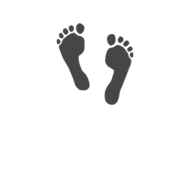}\\
& sleep(s) & \includegraphics[align=c, width=2em]{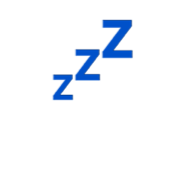} \\
& run(s) fast & \includegraphics[align=c, width=2em]{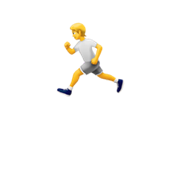}\\
& wave(s) its hand & \includegraphics[align=c, width=2em]{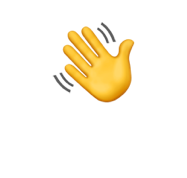}\\
& write(s) a text & \includegraphics[align=c, width=2em]{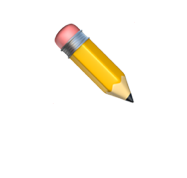}\\
& take(s) a bus & \includegraphics[align=c, width=2em]{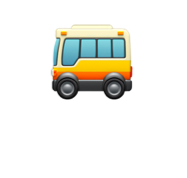}\\
& take(s) a photo & \includegraphics[align=c, width=2em]{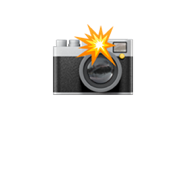}\\
\vspace{1em}
& play(s) soccer & \includegraphics[align=c, width=1em]{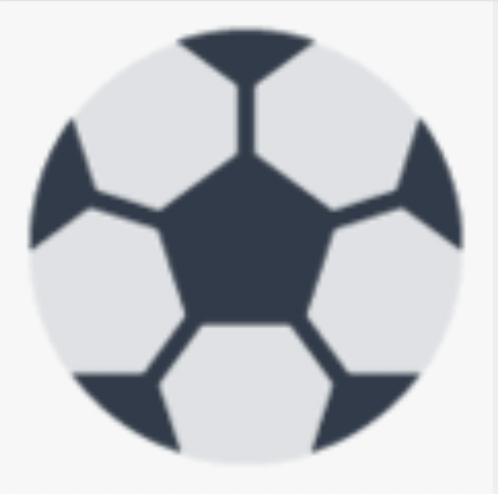}\\
& play(s) baseball & \includegraphics[align=c, width=1em]{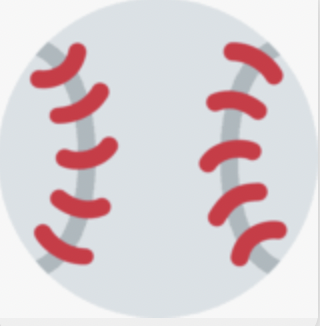}\\
& throw(s) an arrow at a target & \includegraphics[align=c, width=2em]{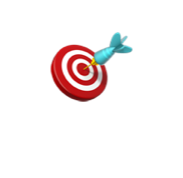}\\
\bottomrule
\end{tabular}
\caption{Vocabularies and their corresponding visual features used in the \textsc{Artificial} dataset.
}
\label{tbl:data}
}
\end{table}

%% file: tables/blimp.tex
\begin{table*}[t]
    \centering
    \renewcommand{\arraystretch}{1.0}
    \tabcolsep  4pt
    \begin{tabular}{@{}ccp{0.62cm}p{0.62cm}p{0.62cm}p{0.62cm}p{0.62cm}p{0.62cm}p{0.62cm}p{0.62cm}p{0.62cm}p{0.62cm}p{0.62cm}p{0.62cm}p{0.62cm}p{0.62cm}p{0.31cm}@{}}
    \toprule
        Vision & \rotatebox{45}{\textsc{Overall}} & \rotatebox{45}{\textsc{Ana. agr}} & \rotatebox{45}{\textsc{D-n agr}} & \rotatebox{45}{\textsc{Irregular}} & \rotatebox{45}{\textsc{S-v agr}} & \rotatebox{45}{\textsc{Arg. str}} & \rotatebox{45}{\textsc{Ellipsis}} & \rotatebox{45}{\textsc{Filler-gap}} & \rotatebox{45}{\textsc{Island}} & \rotatebox{45}{\textsc{NPI}} & \rotatebox{45}{\textsc{Quantifiers}} & \rotatebox{45}{\textsc{Binding}} & \rotatebox{45}{\textsc{Ctrl. Rais.}} & \\
        \cmidrule(lr){1-1} \cmidrule(lr){2-2} \cmidrule(lr){3-15}
        \sight & \textbf{59.1} & 60.3 & \textbf{61.1} & \textbf{70.8} & 61.1 & 61.0 & \textbf{48.9} & \textbf{65.7} & 56.5 & \textbf{43.2} & 72.6 & 61.4 & 59.1 & \\
        \blind & \textbf{59.1} & \textbf{60.5} & 60.3 & 69.3 & 62.0 & 61.6 & \textbf{48.9} & 65.3 & 55.3 & 42.5 & \textbf{73.7} & 62.9 & 59.5 & \\
        \cyclone & 58.8 & 59.4 & 60.4 & 70.7 & \textbf{62.3} & \textbf{62.7} & 42.4 & \textbf{65.7} & \textbf{59.7} & 42.5 & 69.8 & \textbf{66.0} & \textbf{61.1} & \\
        \bottomrule
         
    \end{tabular}
    \caption{Accuracy on each circuit on the BLiMP benchmark. The \sight\ model corresponds to the Vit-base model used in the main experiment, the \blind\ model corresponds to the model trained with a white noise image, and the \cyclone\ model corresponds to the model trained with shuffled image-caption data.}
    \label{tbl:blimp}
\end{table*}

%% file: tables/parameters.tex
\begin{table*}
    \centering
    \begin{tabular}{p{6cm}p{8cm}} \toprule
     \multirow{1}{2cm}{Decoder} & Following the settings in \url{https://huggingface.co/gpt2/blob/main/config.json} \\
    \cmidrule(lr){1-1} \cmidrule(lr){2-2}
    Dropout rate in encoder & 0.1 (attention and hidden state) \\
    \cmidrule(lr){1-1} \cmidrule(lr){2-2}
    Optimizer & AdamW~\cite{loshchilov2018decoupled} \\
    $\;\;\;\;$learning rate & 1e-4 \\
    $\;\;\;\;$betas & (0.9, 0.999) \\
    $\;\;\;\;$epsilon & 1e-8 \\
    \cmidrule(lr){1-1} \cmidrule(lr){2-2}
    Learning scheduler & linear decay \\
    $\;\;\;\;$max steps & 10,000 (\textsc{Natural} setting), 1000 (\textsc{Artificial} setting) \\
    $\;\;\;\;$warm up steps & 0 \\
    $\;\;\;\;$weight decay & 0 \\
    \cmidrule(lr){1-1} \cmidrule(lr){2-2}
    Batchsize & 512 \\
    \cmidrule(lr){1-1} \cmidrule(lr){2-2}
    Beam size & 4 (when computing ROUGE) \\
  \bottomrule
\end{tabular}
    \caption{Common hyperparameters across the models with different vision encoders.}
    \label{tbl:hyperparam}
\end{table*}

%% file: tables/rouge.tex
{\setlength{\fboxsep}{1pt}
\begin{table*}[t]
\centering
\tabcolsep  4pt
\begin{center}
\begin{tabular}{p{2cm}crrrrrr} \toprule
&& \multicolumn{3}{c}{\textsc{Natural}}  & \multicolumn{2}{c}{\textsc{Artificial}} & ImageNet  \\
\cmidrule(lr){3-5} \cmidrule(l){6-7} \cmidrule(l){8-8}
& & \multicolumn{3}{c}{ROUGE-L F1} & \multicolumn{2}{c}{ROUGE-L F1} &  \\
 Models & Vis. & 1,000 & 5,000 & 10,000 & 100 & 500 & Acc@1 \\
\cmidrule(lr){1-1} \cmidrule(lr){2-2} \cmidrule(lr){3-5} \cmidrule(lr){6-7} \cmidrule(l){8-8}
\multirow{2}{2cm}{Vit-base (86M)} & \sighted & $32.0$ & $35.5$ & $37.8$ & $80.5$ & $100.0$ & 84.0 \\
& \diff & $+17.3$ & $+20.2$ & $+22.8$ & $+45.1$ & $+64.5$ & \\
\cmidrule(lr){1-1} \cmidrule(lr){2-2} \cmidrule(lr){3-5} \cmidrule(lr){6-7} \cmidrule(l){8-8}
\multirow{2}{2cm}{Vit-large (307M)}& \sighted & $30.8$ & $35.1$ & $37.9$ & $76.3$ & $100.0$ & 85.2 \\
& \diff & $+16.1$ & $+20.2$ & $+22.6$ & $+40.7$ & $+64.5$  \\
\cmidrule(lr){1-1} \cmidrule(lr){2-2} \cmidrule(lr){3-5} \cmidrule(lr){6-7} \cmidrule(l){8-8}
\multirow{2}{2cm}{Vit-huge (632M)}& \sighted & $29.2$ & $34.1$ & $35.8$ & $59.1$ & $100.0$ & 85.1 \\
& \diff & $+14.9$ & $+18.8$& $+20.5$ & $+23.8$ & $+63.9$ \\
\cmidrule(lr){1-1} \cmidrule(lr){2-2} \cmidrule(lr){3-5} \cmidrule(lr){6-7} \cmidrule(l){8-8}
\multirow{2}{2cm}{Beit-base (86M)}& \sighted & $31.7$ & $34.5$ & $37.4$ & $51.5$ & $100.0$ & 85.2 \\
& \diff & $+15.9$ & $+19.2$ & $+22.1$ & $+16.5$ & $+64.6$ \\
\cmidrule(lr){1-1} \cmidrule(lr){2-2} \cmidrule(lr){3-5} \cmidrule(lr){6-7} \cmidrule(l){8-8}
\multirow{2}{2cm}{Beit-large (307M)}& \sighted & $30.4$ & $37.0$ & $40.2$ & $81.2$ & $100.0$ & 87.4 \\
& \diff & $+15.7$ & $+21.8$ & $+24.9$ & $+46.0$ & $+64.8$ \\
\cmidrule(lr){1-1} \cmidrule(lr){2-2} \cmidrule(lr){3-5} \cmidrule(lr){6-7} \cmidrule(l){8-8}
\multirow{2}{2cm}{Deit-base (86M)}& \sighted & $32.2$ & $35.6$ & $38.2$ & $98.5$ & $100.0$ & 83.4 \\
& \diff & $+18.5$ & $+20.4$ & $+22.9$ & $+63.0$ & $+64.4$ \\
\cmidrule(lr){1-1} \cmidrule(lr){2-2} \cmidrule(lr){3-5} \cmidrule(lr){6-7} \cmidrule(l){8-8}
\multirow{2}{2cm}{Deit-small (22M)}& \sighted & $31.0$ & $34.6$ & $36.6$ & $83.0$ & $100.0$ & 81.2 \\
& \diff & $+16.3$ & $+19.6$ & $+21.2$ & $+47.7$ & $+64.6$ \\
\cmidrule(lr){1-1} \cmidrule(lr){2-2} \cmidrule(lr){3-5} \cmidrule(lr){6-7} \cmidrule(l){8-8}
\multirow{2}{2cm}{Deit-tiny (5M)}& \sighted & $30.1$ & $33.7$ & $35.4$ & $93.2$ & $100.0$ & 74.5 \\
& \diff & $+15.4$ & $+18.4$ & $+20.1$ & $+58.1$ & $+64.6$ \\
\cmidrule(lr){1-1} \cmidrule(lr){2-2} \cmidrule(lr){3-5} \cmidrule(lr){6-7} \cmidrule(l){8-8}
\multirow{2}{2cm}{Swin-base (88M)}& \sighted & $34.3$ & $37.6$ & $40.7$ & $99.3$ & $100.0$ & 85.2 \\
& \diff & $+19.6$ & $+22.3$ & $+25.4$ & $+64.0$ & $+64.3$ \\
\cmidrule(lr){1-1} \cmidrule(lr){2-2} \cmidrule(lr){3-5} \cmidrule(lr){6-7} \cmidrule(l){8-8}
\multirow{2}{2cm}{Swin-large (197M)}& \sighted & $34.5$ & $38.3$ & $41.7$ & $97.6$ & $100.0$ & 87.3 \\
& \diff & $+19.2$ & $+23.4$ & $+26.4$ & $+62.3$ & $+64.3$ \\
\midrule
\multirow{2}{2cm}{Scratch (86M)}& \sighted & $13.94$ & $23.7$ & $24.5$ & $37.3$ & $65.6$ & - \\
& \diff & $+0.16$ & $+8.78$ & $+8.93$ & $+1.88$ & $30.3$ \\
\cmidrule(lr){1-1} \cmidrule(lr){2-2} \cmidrule(lr){3-5} \cmidrule(lr){6-7} \cmidrule(l){8-8}
\multirow{2}{2cm}{Vit-GPT2 (86M)}& \sighted & $32.4$ & $35.3$ & $37.4$ & $93.3$ & $100.0$ & 84.0 \\
& \diff & $+17.7$ & $+20.4$ & $+22.1$ & $+57.7$ & $+64.2$ \\
\bottomrule
\end{tabular}
\end{center}
\caption{ROUGE-L F1 scores of the models at several checkpoints with different training steps. The scores are multiplied by 100. ImageNet accuracy scores are obtained from their original papers.
}
\label{tbl:rouge}
\end{table*}
}

%% file: coling_latex.bbl
\begin{thebibliography}{58}
\providecommand{\natexlab}[1]{#1}

\bibitem[{Alayrac et~al.(2022)Alayrac, Donahue, Luc, Miech, Barr, Hasson, Lenc, Mensch, Millican, Reynolds et~al.}]{Alayrac2022-vo}
Jean-Baptiste Alayrac, Jeff Donahue, Pauline Luc, Antoine Miech, Iain Barr, Yana Hasson, Karel Lenc, Arthur Mensch, Katherine Millican, Malcolm Reynolds, et~al. 2022.
\newblock \href {https://proceedings.neurips.cc/paper_files/paper/2022/file/960a172bc7fbf0177ccccbb411a7d800-Paper-Conference.pdf} {Flamingo: a visual language model for few-shot learning}.
\newblock \emph{Proceedings of NeurIPS 2022}, 35:23716--23736.

\bibitem[{Bao et~al.(2022)Bao, Dong, Piao, and Wei}]{Bao2022-ha}
Hangbo Bao, Li~Dong, Songhao Piao, and Furu Wei. 2022.
\newblock \href {https://openreview.net/forum?id=p-BhZSz59o4} {{BE}it: {BERT} pre-training of image transformers}.
\newblock In \emph{Proceedings of ICLR 2022}.

\bibitem[{Barsalou(2008)}]{Barsalou2008-fl}
Lawrence~W Barsalou. 2008.
\newblock \href {https://www.annualreviews.org/content/journals/10.1146/annurev.psych.59.103006.093639} {Grounded cognition}.
\newblock \emph{Annu. Rev. Psychol.}, 59(1):617--645.

\bibitem[{Bender and Koller(2020)}]{Bender2020-ds}
Emily~M Bender and Alexander Koller. 2020.
\newblock \href {https://aclanthology.org/2020.acl-main.463/} {Climbing towards {{NLU}}: {On} meaning, form, and understanding in the age of data}.
\newblock In \emph{Proceedings of ACL 2020}, pages 5185--5198.

\bibitem[{Brown et~al.(2020)Brown, Mann, Ryder, Subbiah, Kaplan, Dhariwal, Neelakantan, Shyam, Sastry, Askell, Agarwal, Herbert-Voss, Krueger, Henighan, Child, Ramesh, Ziegler, Wu, Winter, Hesse, Chen, Sigler, Litwin, Gray, Chess, Clark, Berner, McCandlish, Radford, Sutskever, and Amodei}]{Brown2020-zt}
Tom~B Brown, Benjamin Mann, Nick Ryder, Melanie Subbiah, Jared Kaplan, Prafulla Dhariwal, Arvind Neelakantan, Pranav Shyam, Girish Sastry, Amanda Askell, Sandhini Agarwal, Ariel Herbert-Voss, Gretchen Krueger, Tom Henighan, Rewon Child, Aditya Ramesh, Daniel~M Ziegler, Jeffrey Wu, Clemens Winter, Christopher Hesse, Mark Chen, Eric Sigler, Mateusz Litwin, Scott Gray, Benjamin Chess, Jack Clark, Christopher Berner, Sam McCandlish, Alec Radford, Ilya Sutskever, and Dario Amodei. 2020.
\newblock \href {https://splab.sdu.edu.cn/GPT3.pdf} {{Language Models are {Few-Shot} Learners}}.
\newblock In \emph{Proceedings of {NeurIPS} 2020}.

\bibitem[{Chomsky(1964)}]{chomsky1964aspects}
Noam Chomsky. 1964.
\newblock \href {https://mitpress.mit.edu/9780262030113/} {Aspects of the theory of syntax}.
\newblock Technical report, MASSACHUSETTS INST OF TECH CAMBRIDGE RESEARCH LAB OF ELECTRONICS.

\bibitem[{Chomsky(1980)}]{Chomsky1980-vx}
Noam Chomsky. 1980.
\newblock Rules and representations.
\newblock \emph{Behavioral and Brain Sciences}, 3(1):1--15.

\bibitem[{Choshen et~al.(2024)Choshen, Cotterell, Hu, Linzen, Mueller, Ross, Warstadt, Wilcox, Williams, and Zhuang}]{Choshen2024-sn}
Leshem Choshen, Ryan Cotterell, Michael~Y Hu, Tal Linzen, Aaron Mueller, Candace Ross, Alex Warstadt, Ethan Wilcox, Adina Williams, and Chengxu Zhuang. 2024.
\newblock \href {https://arxiv.org/abs/2404.06214} {[call for papers] the {2nd} {BabyLM} challenge: Sample-efficient pretraining on a developmentally plausible corpus}.
\newblock \emph{arXiv [cs.CL]}.

\bibitem[{Crain and Nakayama(1987)}]{Crain1987-ji}
Stephen Crain and Mineharu Nakayama. 1987.
\newblock \href {https://tallinzen.net/media/readings/crain_nakayama_1987.pdf} {Structure dependence in grammar formation}.
\newblock \emph{Language}, 63(3):522--543.

\bibitem[{Cubuk et~al.(2020)Cubuk, Zoph, Shlens, and Le}]{Cubuk2020-nt}
Ekin~Dogus Cubuk, Barret Zoph, Jon Shlens, and Quoc Le. 2020.
\newblock \href {https://proceedings.neurips.cc/paper/2020/hash/d85b63ef0ccb114d0a3bb7b7d808028f-Abstract.html} {Randaugment: Practical automated data augmentation with a reduced search space}.
\newblock In \emph{Proceedings of NeurIPS 2020}, volume~33, pages 18613--18624.

\bibitem[{Dosovitskiy et~al.(2021)Dosovitskiy, Beyer, Kolesnikov, Weissenborn, Zhai, Unterthiner, Dehghani, Minderer, Heigold, Gelly, Uszkoreit, and Houlsby}]{Dosovitskiy2020-cu}
Alexey Dosovitskiy, Lucas Beyer, Alexander Kolesnikov, Dirk Weissenborn, Xiaohua Zhai, Thomas Unterthiner, Mostafa Dehghani, Matthias Minderer, Georg Heigold, Sylvain Gelly, Jakob Uszkoreit, and Neil Houlsby. 2021.
\newblock \href {https://openreview.net/forum?id=YicbFdNTTy} {An image is worth 16x16 words: Transformers for image recognition at scale}.
\newblock In \emph{Proceedings of ICLR 2021}.

\bibitem[{Dupoux(2018)}]{Dupoux2016-ia}
Emmanuel Dupoux. 2018.
\newblock \href {https://www.sciencedirect.com/science/article/abs/pii/S0010027717303013} {Cognitive science in the era of artificial intelligence: A roadmap for reverse-engineering the infant language-learner}.
\newblock \emph{Cognition}, 173:43--59.

\bibitem[{Elman(1990)}]{Elman1990-on}
Jeffrey~L Elman. 1990.
\newblock Finding structure in time.
\newblock \emph{Cogn. Sci.}, 14(2):179--211.

\bibitem[{Gilkerson et~al.(2017)Gilkerson, Richards, Warren, Montgomery, Greenwood, Kimbrough~Oller, Hansen, and Paul}]{Gilkerson2017-iz}
Jill Gilkerson, Jeffrey~A Richards, Steven~F Warren, Judith~K Montgomery, Charles~R Greenwood, D~Kimbrough~Oller, John H~L Hansen, and Terrance~D Paul. 2017.
\newblock \href {https://pubs.asha.org/doi/full/10.1044/2016_AJSLP-15-0169} {Mapping the early language environment using all-day recordings and automated analysis}.
\newblock \emph{Am. J. Speech. Lang. Pathol.}, 26(2):248--265.

\bibitem[{Gleitman and Gleitman(1992)}]{Gleitman1992-jo}
Lila~R Gleitman and Henry Gleitman. 1992.
\newblock \href {https://journals.sagepub.com/doi/abs/10.1111/1467-8721.ep10767853} {A picture is worth a thousand words, but that's the problem: The role of syntax in vocabulary acquisition}.
\newblock \emph{Current Directions in Psychological Science}, 1(1):31--35.

\bibitem[{Goldberg(2005)}]{Goldberg2005-jc}
Adele Goldberg. 2005.
\newblock \href {http://llt.cbs.polyu.edu.hk/static/upload/cv/Adele_Goldberg_Constructions_at_Work_The_NatureBookFi.org.PDF} {\emph{Constructions at Work: The Nature of Generalization in Language}}.
\newblock Walter de Gruyter GmbH \& Co. KG.

\bibitem[{Hill and Wagovich(2020)}]{Hill2020-ye}
Margaret~S Hill and Stacy~A Wagovich. 2020.
\newblock \href {https://pubmed.ncbi.nlm.nih.gov/32252839/} {Word learning from context in school-age children: relations with language ability and executive function}.
\newblock \emph{J. Child Lang.}, 47(5):1006--1029.

\bibitem[{Honnibal et~al.(2020)Honnibal, Montani, Van~Landeghem, and Boyd}]{spacy}
Matthew Honnibal, Ines Montani, Sofie Van~Landeghem, and Adriane Boyd. 2020.
\newblock \href {10.5281/zenodo.1212303} {spacy: Industrial-strength natural language processing in python}.

\bibitem[{Johnson et~al.(2012)Johnson, Demuth, and Frank}]{Johnson2012-ba}
Mark Johnson, Katherine Demuth, and Michael Frank. 2012.
\newblock \href {https://aclanthology.org/P12-1093/} {Exploiting social information in grounded language learning via grammatical reduction}.
\newblock In \emph{Proceedings of ACL 2012}, pages 883--891.

\bibitem[{Kirov and Cotterell(2018)}]{kirov-cotterell-2018-recurrent}
Christo Kirov and Ryan Cotterell. 2018.
\newblock \href {https://doi.org/10.1162/tacl_a_00247} {Recurrent neural networks in linguistic theory: Revisiting pinker and prince (1988) and the past tense debate}.
\newblock \emph{TACL}, 6:651--665.

\bibitem[{Kojima et~al.(2020)Kojima, Averbuch-Elor, Rush, and Artzi}]{Kojima2020-ma}
Noriyuki Kojima, Hadar Averbuch-Elor, Alexander Rush, and Yoav Artzi. 2020.
\newblock \href {https://aclanthology.org/2020.acl-main.234} {What is learned in visually grounded neural syntax acquisition}.
\newblock In \emph{Proceedings of ACL 2020}, pages 2615--2635.

\bibitem[{Legate and Yang(2002)}]{Legate2002-nn}
Julie~Anne Legate and Charles~D Yang. 2002.
\newblock \href {https://www.degruyter.com/document/doi/10.1515/tlir.19.1-2.151/html} {Empirical re-assessment of stimulus poverty arguments}.
\newblock \emph{The Linguistic Review}, 19(1-2):151--162.

\bibitem[{{Liu} et~al.(2021){Liu}, {Lin}, {Cao}, {Hu}, {Wei}, {Zhang}, {Lin}, and {Guo}}]{Liu2021-qc}
{Liu}, {Lin}, {Cao}, {Hu}, {Wei}, {Zhang}, {Lin}, and {Guo}. 2021.
\newblock \href {https://openaccess.thecvf.com/content/ICCV2021/html/Liu_Swin_Transformer_Hierarchical_Vision_Transformer_Using_Shifted_Windows_ICCV_2021_paper} {Swin transformer: Hierarchical vision transformer using shifted windows}.
\newblock In \emph{Proceedings of ICCV 2021}, pages 9992--10002.

\bibitem[{Loshchilov and Hutter(2018)}]{loshchilov2018decoupled}
Ilya Loshchilov and Frank Hutter. 2018.
\newblock \href {https://openreview.net/forum?id=Bkg6RiCqY7} {Decoupled weight decay regularization}.
\newblock In \emph{Proceedings of ICLR 2018}.

\bibitem[{Ma et~al.(2021)Ma, Shen, Yoshikawa, Iwakura, Beck, and Baldwin}]{Ma2021-dk}
Chunpeng Ma, Aili Shen, Hiyori Yoshikawa, Tomoya Iwakura, Daniel Beck, and Timothy Baldwin. 2021.
\newblock \href {https://aclanthology.org/2021.eacl-main.4/} {On the (in)effectiveness of images for text classification}.
\newblock In \emph{Proceedings of EACL 2021}, pages 42--48.

\bibitem[{Ma et~al.(2023)Ma, Pan, and Chai}]{Ma2023-tv}
Ziqiao Ma, Jiayi Pan, and Joyce Chai. 2023.
\newblock \href {https://aclanthology.org/2023.acl-long.31/} {World-to-words: Grounded open vocabulary acquisition through fast mapping in vision-language models}.
\newblock In \emph{Proceedings of ACL 2023}, pages 524--544.

\bibitem[{McCoy et~al.(2018)McCoy, Frank, and Linzen}]{mccoy2018revisiting}
R~Thomas McCoy, Robert Frank, and Tal Linzen. 2018.
\newblock \href {https://escholarship.org/uc/item/12w6n3xn} {Revisiting the poverty of the stimulus: hierarchical generalization without a hierarchical bias in recurrent neural networks}.
\newblock In \emph{40th Annual Meeting of the Cognitive Science Society: Changing Minds, CogSci 2018}, pages 2096--2101.

\bibitem[{McCoy et~al.(2020)McCoy, Frank, and Linzen}]{McCoy2020-mc}
R~Thomas McCoy, Robert Frank, and Tal Linzen. 2020.
\newblock \href {https://aclanthology.org/2020.tacl-1.9/} {Does syntax need to grow on trees? sources of hierarchical inductive bias in {Sequence-to-Sequence} networks}.
\newblock \emph{TACL}, 8:125--140.

\bibitem[{McCoy et~al.(2019)McCoy, Pavlick, and Linzen}]{McCoy2019-ek}
Tom McCoy, Ellie Pavlick, and Tal Linzen. 2019.
\newblock \href {https://aclanthology.org/P19-1334/} {Right for the wrong reasons: Diagnosing syntactic heuristics in natural language inference}.
\newblock In \emph{Proceedings of ACL 2019}, pages 3428--3448.

\bibitem[{McDonough et~al.(2011)McDonough, Song, Hirsh-Pasek, Golinkoff, and Lannon}]{McDonough2011-oh}
Colleen McDonough, Lulu Song, Kathy Hirsh-Pasek, Roberta~Michnick Golinkoff, and Robert Lannon. 2011.
\newblock \href {https://www.ncbi.nlm.nih.gov/pmc/articles/PMC3043374/} {An image is worth a thousand words: why nouns tend to dominate verbs in early word learning}.
\newblock \emph{Dev. Sci.}, 14(2):181--189.

\bibitem[{Mitchell(1980)}]{Mitchell1980-nj}
Tom~M Mitchell. 1980.
\newblock \href {https://www.cs.cmu.edu/~tom/pubs/NeedForBias_1980.pdf} {\emph{The need for biases in learning generalizations}}.
\newblock Citeseer.

\bibitem[{Mueller et~al.(2022)Mueller, Frank, Linzen, Wang, and Schuster}]{Mueller2022-gt}
Aaron Mueller, Robert Frank, Tal Linzen, Luheng Wang, and Sebastian Schuster. 2022.
\newblock \href {https://doi.org/10.18653/v1/2022.findings-acl.106} {Coloring the blank slate: Pre-training imparts a hierarchical inductive bias to sequence-to-sequence models}.
\newblock In \emph{Findings of ACL 2022}, pages 1352--1368.

\bibitem[{Nikolaus et~al.(2019)Nikolaus, Abdou, Lamm, Aralikatte, and Elliott}]{Nikolaus2019-ay}
Mitja Nikolaus, Mostafa Abdou, Matthew Lamm, Rahul Aralikatte, and Desmond Elliott. 2019.
\newblock \href {https://aclanthology.org/K19-1009/} {Compositional generalization in image captioning}.
\newblock In \emph{Proceedings of CoNLL 2019}, pages 87--98.

\bibitem[{OpenAI(2023)}]{achiam2023gpt}
OpenAI. 2023.
\newblock \href {https://cdn.openai.com/papers/gpt-4.pdf} {Gpt-4 technical report}.
\newblock Technical report, OpenAI.

\bibitem[{Perfors et~al.(2011)Perfors, Tenenbaum, and Regier}]{Perfors2011-mu}
Amy Perfors, Joshua~B Tenenbaum, and Terry Regier. 2011.
\newblock \href {https://www.sciencedirect.com/science/article/abs/pii/S0010027710002593} {The learnability of abstract syntactic principles}.
\newblock \emph{Cognition}, 118(3):306--338.

\bibitem[{Perkins and Lidz(2021)}]{Perkins2021-jy}
Laurel Perkins and Jeffrey Lidz. 2021.
\newblock \href {https://www.pnas.org/doi/full/10.1073/pnas.2026469118} {Eighteen-month-old infants represent nonlocal syntactic dependencies}.
\newblock \emph{Proceedings of the National Academy of Sciences}, 118(41):e2026469118.

\bibitem[{Qin et~al.(2024)Qin, Wang, and Lake}]{Qin2024-ng}
Yulu Qin, Wentao Wang, and Brenden~M Lake. 2024.
\newblock A systematic investigation of learnability from single child linguistic input.
\newblock \emph{arXiv [cs.CL]}.

\bibitem[{Qu and Chai(2008)}]{Qu2008-fx}
Shaolin Qu and Joyce Chai. 2008.
\newblock \href {https://aclanthology.org/D08-1026/} {Incorporating temporal and semantic information with eye gaze for automatic word acquisition in multimodal conversational systems}.
\newblock In \emph{Proceedings of EMNLP 2008}, pages 244--253.

\bibitem[{Radford et~al.(2021)Radford, Kim, Hallacy, Ramesh, Goh, Agarwal, Sastry, Askell, Mishkin, Clark et~al.}]{Radford2021-bt}
Alec Radford, Jong~Wook Kim, Chris Hallacy, Aditya Ramesh, Gabriel Goh, Sandhini Agarwal, Girish Sastry, Amanda Askell, Pamela Mishkin, Jack Clark, et~al. 2021.
\newblock \href {https://proceedings.mlr.press/v139/radford21a/radford21a.pdf} {Learning transferable visual models from natural language supervision}.
\newblock In \emph{Proceedings of ICML}, pages 8748--8763.

\bibitem[{Radford et~al.(2019)Radford, Wu, Child, Luan, Amodei, and Sutskever}]{Radford_undated-nn}
Alec Radford, Jeffrey Wu, Rewon Child, David Luan, Dario Amodei, and Ilya Sutskever. 2019.
\newblock \href {https://www.semanticscholar.org/paper/Language-Models-are-Unsupervised-Multitask-Learners-Radford-Wu/9405cc0d6169988371b2755e573cc28650d14dfe} {{Language Models are Unsupervised Multitask Learners}}.
\newblock Technical report, OpenAI.

\bibitem[{Ramesh et~al.(2021)Ramesh, Pavlov, Goh, Gray, Voss, Radford, Chen, and Sutskever}]{Ramesh2021-sy}
Aditya Ramesh, Mikhail Pavlov, Gabriel Goh, Scott Gray, Chelsea Voss, Alec Radford, Mark Chen, and Ilya Sutskever. 2021.
\newblock \href {https://proceedings.mlr.press/v139/ramesh21a.html} {Zero-shot text-to-image generation}.
\newblock In \emph{Proceedings of ICML 2021}, volume 139, pages 8821--8831.

\bibitem[{R{\"a}s{\"a}nen and Khorrami(2019)}]{rasanen2019computational}
Okko R{\"a}s{\"a}nen and Khazar Khorrami. 2019.
\newblock \href {https://www.isca-archive.org/interspeech_2019/rasanen19_interspeech.html} {A computational model of early language acquisition from audiovisual experiences of young infants}.
\newblock In \emph{Interspeech}, pages 3594--3598. International Speech Communication Association ISCA.

\bibitem[{Roy and Reiter(2005)}]{Roy2005-wo}
Deb Roy and Ehud Reiter. 2005.
\newblock \href {https://www.sciencedirect.com/science/article/pii/S0004370205001049} {Connecting language to the world}.
\newblock \emph{Artificial Intelligence}, 167(1-2):1--12.

\bibitem[{Sharma et~al.(2018)Sharma, Ding, Goodman, and Soricut}]{Sharma2018-wl}
Piyush Sharma, Nan Ding, Sebastian Goodman, and Radu Soricut. 2018.
\newblock \href {https://aclanthology.org/P18-1238/} {Conceptual captions: A cleaned, hypernymed, image alt-text dataset for automatic image captioning}.
\newblock In \emph{Proceedings of ACL 2018}, pages 2556--2565.

\bibitem[{Shi et~al.(2019)Shi, Mao, Gimpel, and Livescu}]{Shi2019-ot}
Haoyue Shi, Jiayuan Mao, Kevin Gimpel, and Karen Livescu. 2019.
\newblock \href {https://aclanthology.org/P19-1180/} {Visually grounded neural syntax acquisition}.
\newblock In \emph{Proceedings of ACL 2019}, pages 1842--1861.

\bibitem[{Spector(2007)}]{Spector2007-gb}
Benjamin Spector. 2007.
\newblock \href {https://link.springer.com/chapter/10.1057/9780230210752_9} {Aspects of the pragmatics of plural morphology: On {Higher-Order} implicatures}.
\newblock \emph{Presupposition and Implicature in Compositional Semantics}, pages 243--281.

\bibitem[{Touvron et~al.(2021)Touvron, Cord, Douze, Massa, Sablayrolles, and Jegou}]{Touvron2021-at}
Hugo Touvron, Matthieu Cord, Matthijs Douze, Francisco Massa, Alexandre Sablayrolles, and Herve Jegou. 2021.
\newblock \href {https://proceedings.mlr.press/v139/touvron21a.html} {Training data-efficient image transformers \& distillation through attention}.
\newblock In \emph{Proceedings of ICML 2021}, volume 139, pages 10347--10357.

\bibitem[{Vong et~al.(2024)Vong, Wang, Orhan, and Lake}]{Vong2024-ro}
Wai~Keen Vong, Wentao Wang, A~Emin Orhan, and Brenden~M Lake. 2024.
\newblock \href {https://www.science.org/doi/10.1126/science.adi1374} {Grounded language acquisition through the eyes and ears of a single child}.
\newblock \emph{Science}, 383(6682):504--511.

\bibitem[{Wang et~al.(2023)Wang, Vong, Kim, and Lake}]{Wang2023-pb}
Wentao Wang, Wai~Keen Vong, Najoung Kim, and Brenden~M Lake. 2023.
\newblock \href {https://onlinelibrary.wiley.com/doi/full/10.1111/cogs.13305} {Finding structure in one child's linguistic experience}.
\newblock \emph{Cogn. Sci.}, 47(6):e13305.

\bibitem[{Warstadt and Bowman(2020)}]{Warstadt2020-tk}
Alex Warstadt and Samuel~R Bowman. 2020.
\newblock \href {https://cognitivesciencesociety.org/cogsci20/papers/0381/0381.pdf} {Can neural networks acquire a structural bias from raw linguistic data?}
\newblock In \emph{Proceedings of Cogsci}, pages 1737--1743.

\bibitem[{Warstadt and Bowman(2022)}]{Warstadt2022-zi}
Alex Warstadt and Samuel~R Bowman. 2022.
\newblock \href {https://arxiv.org/abs/2208.07998} {What artificial neural networks can tell us about human language acquisition}.
\newblock \emph{Algebraic Structures in Natural Language}.

\bibitem[{Warstadt et~al.(2020{\natexlab{a}})Warstadt, Parrish, Liu, Mohananey, Peng, Wang, and Bowman}]{Warstadt2020-or}
Alex Warstadt, Alicia Parrish, Haokun Liu, Anhad Mohananey, Wei Peng, Sheng-Fu Wang, and Samuel Bowman. 2020{\natexlab{a}}.
\newblock \href {https://aclanthology.org/2020.tacl-1.25/} {{BLiMP}: The benchmark of linguistic minimal pairs for english}.
\newblock \emph{TACL}, 8:377--392.

\bibitem[{Warstadt et~al.(2020{\natexlab{b}})Warstadt, Zhang, Li, Liu, and Bowman}]{Warstadt2020-wo}
Alex Warstadt, Yian Zhang, Xiaocheng Li, Haokun Liu, and Samuel~R Bowman. 2020{\natexlab{b}}.
\newblock \href {https://aclanthology.org/2020.emnlp-main.16/} {Learning which features matter: {{R}o{BERT}a} acquires a preference for linguistic generalizations (eventually)}.
\newblock In \emph{Proceedings of EMNLP}, pages 217--235, Online.

\bibitem[{Wilson(2006)}]{Wilson2006-qo}
Colin Wilson. 2006.
\newblock \href {https://onlinelibrary.wiley.com/doi/abs/10.1207/s15516709cog0000_89} {Learning phonology with substantive bias: an experimental and computational study of velar palatalization}.
\newblock \emph{Cogn. Sci.}, 30(5):945--982.

\bibitem[{Yedetore et~al.(2023)Yedetore, Linzen, Frank, and McCoy}]{Yedetore2023-sw}
Aditya Yedetore, Tal Linzen, Robert Frank, and R~Thomas McCoy. 2023.
\newblock \href {https://aclanthology.org/2023.acl-long.521/} {How poor is the stimulus? evaluating hierarchical generalization in neural networks trained on child-directed speech}.
\newblock In \emph{Proceedings of ACL 2023}, pages 9370--9393.

\bibitem[{Yun et~al.(2021)Yun, Sun, and Pavlick}]{Yun2021-wi}
Tian Yun, Chen Sun, and Ellie Pavlick. 2021.
\newblock \href {https://aclanthology.org/2021.findings-emnlp.370/} {Does vision-and-language pretraining improve lexical grounding?}
\newblock In \emph{Proceedings of Findings of EMNLP 2021}, pages 4357--4366.

\bibitem[{Zhuang et~al.(2024)Zhuang, Fedorenko, and Andreas}]{Zhuang2024-fy}
Chengxu Zhuang, Evelina Fedorenko, and Jacob Andreas. 2024.
\newblock \href {https://aclanthology.org/2024.naacl-long.71/} {Visual grounding helps learn word meanings in low-data regimes}.
\newblock In \emph{Proceedings of ACL 2024}, pages 1311--1329.

\bibitem[{Zweig(2009)}]{Zweig2009-ff}
Eytan Zweig. 2009.
\newblock \href {https://link.springer.com/article/10.1007/s10988-009-9064-3} {Number-neutral bare plurals and the multiplicity implicature}.
\newblock \emph{Linguistic Philosophy}, 32(4):353--407.

\end{thebibliography}
